\documentclass[lettersize,journal]{IEEEtran}
\usepackage{amsmath,amsfonts}
\usepackage{amsthm}
\usepackage{algorithmic}
\usepackage{algorithm}
\usepackage{array}

\usepackage{textcomp}
\usepackage{stfloats}
\usepackage{url}
\usepackage{verbatim}
\usepackage{graphicx}
\usepackage{cite}
\usepackage{epsfig}
\usepackage{amssymb}
\usepackage{mathtools}
\usepackage{xcolor}
\usepackage{multirow}
\usepackage{booktabs}
\usepackage{bm}
\usepackage[misc]{ifsym}
\usepackage{graphicx}
\usepackage{amsmath}
\usepackage{amssymb}
\usepackage{booktabs}
\usepackage{multirow}
\usepackage{caption}
\usepackage{subcaption}
\usepackage[normalem]{ulem}
\usepackage{authblk}
\usepackage{hyperref} 
\usepackage{makecell}
\useunder{\uline}{\ul}{}
\begin{document}

\title{Practical No-box Adversarial Attacks with Training-free Hybrid Image Transformation}

% \author{\textbf{
% Qilong Zhang\inst{1}, 
% Youheng Sun\inst{1},
% Chaoning Zhang\inst{3},
% Chaoqun Li\inst{1},
% Xuanhan Wang\inst{2},
% Jingkuan Song\inst{1},
% Lianli Gao*\inst{1}}
% % ,~\IEEEmembership{Senior Member,~IEEE}
% % ,~\IEEEmembership{Member,~IEEE,} \\

% \thanks{*Corresponding author.}
% % \thanks{Manuscript received April 19, 2021; revised August 16, 2021.}
% \institute{Center for Future Media, University of Electronic Science and Technology of China
% \and
% Shenzhen Institute for Advanced Study, University of Electronic Science and Technology of China
% \and
% Kyung Hee University}
% \email{qilong.zhangdl@gmail.com,
% youheng.sun@std.uestc.edu.cn, 
% chaoningzhang1990@gmail.com,
% lichaoqunuestc@gmail.com, 
% wxuanhan@hotmail.com,
% jingkuan.song@gmail.com,
% lianli.gao@uestc.edu.cn}
% }

\author{Qilong Zhang\textsuperscript{1},
 Youheng Sun\textsuperscript{1},
 Chaoning Zhang\textsuperscript{3},
 Chaoqun Li\textsuperscript{1},
 Xuanhan Wang\textsuperscript{2},
 Jingkuan Song\textsuperscript{1},
 Lianli Gao*\textsuperscript{1}\thanks{*Corresponding author}\\
\textsuperscript{1}
Center for Future Media, University of Electronic Science and Technology of China\\
\textsuperscript{2}Shenzhen Institute for Advanced Study, University of Electronic Science and Technology of China\\
\textsuperscript{3} Kyung Hee University\\
{\tt\small \href{mailto:qilong.zhangdl@gmail.com}{qilong.zhangdl@gmail.com},
\href{mailto:youheng.sun@std.uestc.edu.cn}{youheng.sun@std.uestc.edu.cn},
\href{mailto:chaoningzhang1990@gmail.com}{chaoningzhang1990@gmail.com},
\href{mailto:lichaoqunuestc@gmail.com}{lichaoqunuestc@gmail.com},
\href{mailto:wxuanhan@hotmail.com}{wxuanhan@hotmail.com},
\href{mailto:jingkuan.song@gmail.com}{jingkuan.song@gmail.com},
\href{mailto:lianli.gao@uestc.edu.cn}{lianli.gao@uestc.edu.cn}

}}

% ########

\markboth{Journal of \LaTeX\ Class Files,~Vol.~14, No.~8, August~2021}%
{Shell \MakeLowercase{\textit{et al.}}: A Sample Article Using IEEEtran.cls for IEEE Journals}


\maketitle

\begin{abstract}
% This document describes the most common article elements and how to use the IEEEtran class with \LaTeX \ to produce files that are suitable for submission to the IEEE.  IEEEtran can produce conference, journal, and technical note (correspondence) papers with a suitable choice of class options. 

Recently, the adversarial vulnerability of deep neural networks (DNNs) has raised increasing attention. 
Among all the threat models, no-box attacks are the most practical but extremely challenging since they neither rely on any knowledge of the target model or similar substitute model, nor access the dataset for training a new substitute model. Although a recent method has attempted such an attack in a loose sense, its performance is not good enough and computational overhead of training is expensive.
In this paper, we move a step forward and show the existence of a \textbf{training-free} adversarial perturbation under the no-box threat model, which can be successfully used to attack different DNNs in real-time.
Motivated by our observation that high-frequency component (HFC) is dominant in low-level features and plays a crucial role in classification, we attack an image mainly by suppression of the original HFC and adding of noisy HFC.
We empirically and experimentally analyze the requirements of effective noisy HFC and show that it should be regionally homogeneous, repeating and dense.
Remarkably, on ImageNet dataset, our method attacks ten well-known models with a success rate of \textbf{98.13\%} on average, which outperforms state-of-the-art no-box attacks by \textbf{29.39\%}. Furthermore, our method is even competitive to mainstream transfer-based black-box attacks. Our code is available at our Supplementary.

\end{abstract}

\begin{IEEEkeywords}
% Article submission, IEEE, IEEEtran, journal, \LaTeX, paper, template, typesetting.
Adversarial example, No-box, Training-free, Practical.
\end{IEEEkeywords}

\section{Introduction}
% \IEEEPARstart{T}{his} file is intended to serve as a ``sample article file''
% for IEEE journal papers produced under \LaTeX\ using
% IEEEtran.cls version 1.8b and later. The most common elements are covered in the simplified and updated instructions in ``New\_IEEEtran\_how-to.pdf''. For less common elements you can refer back to the original ``IEEEtran\_HOWTO.pdf''. It is assumed that the reader has a basic working knowledge of \LaTeX. Those who are new to \LaTeX \ are encouraged to read Tobias Oetiker's ``The Not So Short Introduction to \LaTeX ,'' available at: \url{http://tug.ctan.org/info/lshort/english/lshort.pdf} which provides an overview of working with \LaTeX.

\IEEEPARstart{D}{eep} neural networks (DNNs)~\cite{vgg,inc-v3,inc-v4,resnet,densenet,wang,10704987} are widely known to be vulnerable to adversarial examples~\cite{TMM4,TMM5,TMM6,TMM7,TMM8,gaker}, \textit{i.e.}, a human-imperceptible perturbation can lead to misclassification. In adversarial machine learning, the term \textit{threat model} defines the rules of the attack, \textit{i.e.}, what resources the attacker can access. Formally, attack scenarios are often divided into the white-box and the black-box. For the white-box threat model~\cite{szegedy2013intriguing,goodfellow2014explaining}, the attacker has full knowledge of a target model, such as model parameters and distribution of training dataset. Recognizing the threat of the white-box, a model owner is unlikely to leak a model's information to public. 
Thus, the white-box attack is often used to evaluate model robustness for revealing its weakest point~\cite{madry2017towards}, but not considered as a practical attack. 
\begin{figure}[h]
	\centering
	\includegraphics[height=2.75cm]{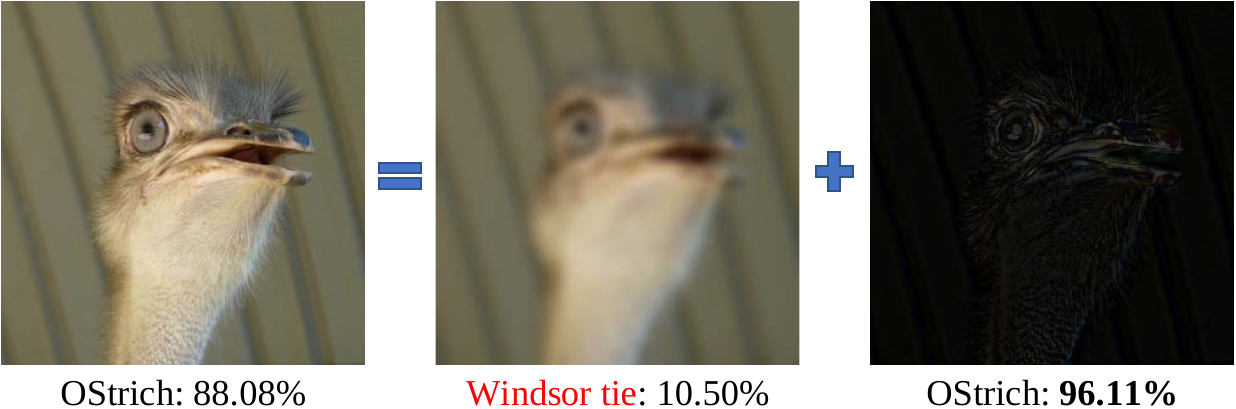}
	\caption{The confidence of a raw image (\textbf{left}), its low-frequency component (\textbf{middle}) and high-frequency component (\textbf{right}) 
	on Inc-v3~\cite{inc-v3}.}
	\label{fig1}
\end{figure}
To remedy it, numerous works have investigated a more realistic scenario, where the attacker does not require full knowledge of the target model, \textit{i.e.}, the backpropagation on the target model is prohibited. This threat model is called query-based black-box attack~\cite{papernot2016transferability,chen2017zoo,yan2019subspace,chen2020hopskipjumpattack}. 
%Under this setting, existing attack methods~\cite{papernot2016transferability,papernot2017practical,tramer2016stealing,zhou2020dast} also make the most of free query to generate synthetic datasets for training a substitute model with which the attacker can apply any white-box attack techniques for generating adversarial examples. Due to the transferability, the resulting adversarial examples are expected to also attack attack the target model~\cite{papernot2016transferability,papernot2017practical}. 
However, this branch usually involves a major concern of being resource-intensive in terms of query cost and time, which inevitably causes suspicion due to repeated queries to the model with almost the same adversarial image.
% . In real-world attack scenarios, even if we ignore such concerns, the query-based black-box attack may still be infeasible because it will cause suspicion due to repeated queries to the model with almost the same adversarial image.
% \textit{e.g.}, the model API is inaccessible to the attacker. Moreover, it might cause suspicion due to repeated queries to the model with almost the same adversarial image. 
To alleviate this issue, another branch of black-box threat model called transfer-based attack~\cite{dong2018boosting,xie2019improving,cda,Zhang2022BIA,ncf} is proposed. In this branch, adversarial examples are crafted via the substitute model which is pre-trained in the same domain of target model. However, transferability heavily depends on how large the gap is between the substitute model and target model. In practice, the gap is large because the target model's structure and training technique are usually not publicly available due to security and privacy concerns.

%\textcolor{red}{For the remainder of this work, unless specified, black-box attack also refers to this threat model.}

%Another variant of black-box threat model~\cite{dong2018boosting,xie2019improving,dong2019evading,sgm,sinifgsm,pifgsm,pifgsm++,ssm}, prohibits the query to the target model for generating the adversarial examples. This is also termed no-box attack~\cite{chen2017zoo} since oracling the target model is fully prohibited. In this work, we call it no-box attack for avoiding confusion with black-box attack that allows query with forward propagation. For the no-box threat model, the whole, or at least partial, training dataset is needed for training a substitute model. In practice, the dataset for training the target model is often not publicly available due to the security and privacy concerns. 
From the discussion above, we argue that both the white-box and black-box attacks can hardly be considered as the most practical attacks. The most practical attack should satisfy two criteria: (a) \textbf{model-free}: no dependence on the pre-trained substitute model or the target model for either backward propagation or only forward query; (b) \textbf{data-free}: no dependence on distribution of training dataset. We term it no-box attack. A recent work~\cite{li2020practical} is the first as well as the only work (to our knowledge) to have attempted such an attack in a loose sense. %Nonetheless, their threat can be seen as data-free no-box attack, but not in a strict sense.
Specifically, their approach attempts to train a substitute model by adopting the classical auto-encoder model. Overall, to attack a certain sample, their approach consists of three steps: (1) collecting a small number of images; (2) training a substitute model; (3) white-box attack on the substitute model. 
% Their threat model still requires a small number of auxiliary samples, such as 20 images.
Admittedly, collecting a small number of auxiliary samples might not be difficult in most cases, but is still infeasible in some security-sensitive applications.
Besides, if a new sample, especially from a different class, needs to be attacked, the above process needs to be repeated. Thus, their approach is resource-intensive. % Besides, their attack success rate is usually lower than existing black-box attacks.  % After their success  Except for the need of collecting auxiliary samples, their approach has the following disadvantages: (a) resource-intensive due to requiring to train a model before attacking a certain sample

By contrast, our approach does not require any of the above steps and is even training-free. With the visualization technique proposed by~\cite{ref_a24}, we observe that the high-frequency component (HFC), \textit{e.g.}, the edge and texture features, is dominant in shallow layers and the low-frequency component (LFC), \textit{e.g.}, the plain areas in the image, is paid less attention to be extracted. Combined with the insight into the classification logic of DNNs in Sec.~\ref{motivation}, we argue that HFC plays a crucial role in recognition. 
As shown in Fig.~\ref{fig1}, without LFC, the confidence of HFC is even higher than the raw image. Although it does not hold true for all samples, it does demonstrate the importance of HFC. 

Motivated by this, we take the idea of hybrid image~\cite{art} and propose a novel \textbf{Hybrid Image Transformation (HIT)} attack method to craft adversarial examples. Formally, it only needs three steps but can effectively fool various DNNs without any training:
First, due to the training-free setting and inspired by the analysis from Sec.~\ref{effectivehfc}, we simply 
% utilize matplotlib\footnote{\label{note1}\url{https://matplotlib.org/}} tool to 
draw several geometric patterns which serve as the proto-patterns, and the resulting synthesized adversarial patches are thus richer in \textbf{regionally homogeneous}, \textbf{repeating} and \textbf{dense} HFC. %In this paper, we test three patterns (see Fig.~\ref{3tu}) to avoid cherry-picking and verify the validity of our approach.
Second, we extract LFC of the raw image and HFC of the adversarial patch. Finally, we combine these two pieces of components and clip them to the $\varepsilon$-ball of the raw image to get the resulting adversarial hybrid example. Extensive experiments on ImageNet demonstrate the effectiveness of our method. By attacking ten state-of-the-art models in the no-box manner, our HIT significantly increases the average success rate from 68.74\% to \textbf{98.13\%}. Notably, our HIT is even competitive to mainstream transfer-based black-box attacks.

% \section{The Design, Intent, and \\ Limitations of the Templates}
% The templates are intended to {\bf{approximate the final look and page length of the articles/papers}}. {\bf{They are NOT intended to be the final produced work that is displayed in print or on IEEEXplore\textsuperscript{\textregistered}}}. They will help to give the authors an approximation of the number of pages that will be in the final version. The structure of the \LaTeX\ files, as designed, enable easy conversion to XML for the composition systems used by the IEEE. The XML files are used to produce the final print/IEEEXplore pdf and then converted to HTML for IEEEXplore.

\section{Related Work}
% \textbf{Adversarial Attack.}
\subsection{Adversarial Attack}
Let $\bm{x}$ denote the raw image, $\bm{x^{adv}}$ and $y$ denote the corresponding adversarial example and true label respectively. 
Follow previous works~\cite{li2020practical,dong2018boosting,Zhang2022BIA}, we use $l_\infty$-norm to measure the perceptibility of adversarial perturbations, \textit{i.e.}, $||\bm{x^{adv}}-\bm{x}||_\infty \leq \varepsilon$. 
In this paper, we focus on non-targeted attacks~\cite{dong2018boosting,xie2019improving,sgm,ghost,sinifgsm,pifgsm,Zhang2022BIA,TMM1,TMM2} which aim to cause misclassification of DNNs $f(\cdot)$, \textit{i.e.}, $f(\bm{x^{adv}})\neq y$. 
 %Since the first discovery of adversarial examples, numerous attack methods have been proposed under different threat models. The white-box threat model assumes full access to the model architectures
 \begin{figure*}[h]
	\centering
	\includegraphics[height=3.8cm]{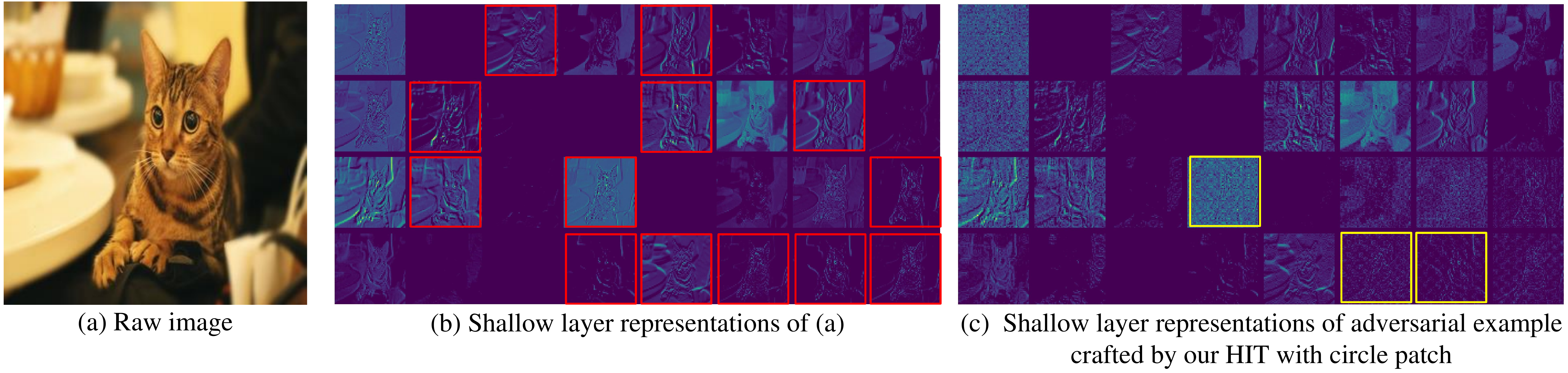}
	\caption{The visualization for the shallow layer (``activation 2”) feature maps of Inception-v3~\cite{inc-v3} w.r.t. the input (a) and its corresponding adversarial example crafted by our HIT.}
	\label{featurespace}
\end{figure*}

% \textbf{Competitors.} 
\subsection{Competitors}
Transferability is an important property for adversarial examples~\cite{papernot2016transferability}. With it, the resulting adversarial example crafted via one model may fool others. \textit{For the traditional black-box threat model},~\cite{goodfellow2014explaining} argue that the vulnerability of DNNs is their linear nature and thus efficiently generate adversarial examples by performing FGSM which is a single-step attack. 
% Papernot \textit{et al.}~\cite{papernot2017practical} train a local model with many queries to substitute for the target model. 
~\cite{dong2018boosting} introduce a momentum term to stabilize the update direction.~\cite{xie2019improving} apply diverse input patterns to improve the transferability of adversarial examples.~\cite{dong2019evading} propose a translation-invariant attack to mitigate the effect of different discriminative regions between models. 
~\cite{pifgsm} introduce patch-wise perturbation to perturb more information in discriminative regions.
~\cite{long2023frequency} propose a novel spectrum transformation and thus generate diverse spectrum saliency maps.
\textit{For the practical black-box threat model}, ~\cite{cda} observe that using dataset from other domain to train generator networks against target domain models can also fool other black-box models from this target domain. ~\cite{dr} propose a dispersion reduction attack to make the low-level features featureless.~\cite{ssp} design a self-supervised perturbation mechanism for enabling a transferable defense approach.~\cite{Zhang2022BIA} train generator networks on just one domain but can fool models from unknown target domains. % Notably, context-aware information has been considered in~\cite{cai2022zero} for attacking object detectors in zero-shot manner but still requires training data to pre-train a substitute model. 
\textit{For the no-box threat model}, ~\cite{li2020practical} attempt to attack the target model without any model query or the accessible pre-trained substitute model. In their work, with a limited amount of data, they try different mechanisms (with or without supervised technique) to train the substitute model, and then utilize this model to craft transferable adversarial examples. 
% A concurrent work~\cite{zhang2021data} has adopted a similar approach but only explores naive patterns like check board. From the data perspective~\cite{li2020practical,zhang2021data}, a recent work~\cite{sun2022towards} reduces the amount of training data but is not fully data-free. 
Unlike these approaches, our method does not depend on transferability since we do not need any substitute model. In this paper, we craft adversarial examples from the perspective of classification logic of DNNs.

\subsection{Frequency Perspective on DNNs}
Our approach is highly inspired by existing works which explain the generalization and adversarial vulnerability of DNNs from the frequency perspective. The fact that DNNs have good generalization while being vulnerable to small adversarial perturbations has motivated~\cite{jo2017measuring,HF} to investigate the underlying mechanism, suggesting that surface-statistical content with high-frequency property is essential for the classification task. From the perspective of texture vs. shape, \cite{texture-bais,HF} reveal that DNNs are biased towards texture instead of shape. Since the texture content is considered to have high-frequency property, their finding can be interpreted as the DNN being biased towards HFC. On the other hand, adversarial perturbations are also known to have the high-frequency property and various defense methods have also been motivated from this insight~\cite{aydemir2018effects,das2018shield,liu2019feature,featuredenoising}. 
Nonetheless, it remains unknown whether manually designed high-frequency patterns are sufficient for attacking the network.

\section{Methodology}
\label{uip}
Although many adversarial attack methods~\cite{dong2018boosting,liu2019universal,li2020practical,Zhang2022BIA,wang2023towards} have achieved impressive results in black-box or no-box cases, they all need training, especially for query-based~\cite{papernot2016transferability,zhou2020dast} and no-box adversarial perturbations~\cite{li2020practical} whose training costs are usually expensive. Then a natural question arises: \textit{Is it possible to generate robust adversarial perturbations without any training?} In the following subsections, we will give our answer and introduce our design.

%\textcolor{red}{In the following sections, we first give an insight into DNNs in Sec.~\ref{fm}, and show that a training-free universal adversarial perturbation is possible. We then analyze what makes a robust perturbation in Sec.~\ref{tuap}. Motivated by these, we finally propose our Universal Image Process (HIT) attack method in Sec.~\ref{ahi} to achieve this goal.}

%\subsection{Rethink the Classification Logic of DNNs}
%\label{rethink}
%In this section, we analyze the classification logic of DNNs from two aspects: Sec.~\ref{fm} gives an insight into intermediate feature space, Sec.~\ref{tuap} empirically analyzes the properties required for robust training-free adversarial perturbations.

\subsection{Motivation}
\label{motivation}

%In this section, we 
%\subsection{Intermediate feature representations}
%Since a lot of efforts have been made in recent years, DNNs demonstrate impressive classification performance on the ImageNet benchmark, \textit{e.g.}, Inception model~\cite{ref_a19,ref_a20} and MobileNet~\cite{mob}. 
%To better explain the behavior of DNNs, many impressive works have been proposed. For example, Zeiler \textit{et al.}~\cite{ref_a24} propose a visualization technique that gives insight into the function of intermediate feature layers. 
%Based on it, we show the result in Fig.~\ref{feature img} and observe that the feature map in the shallow layers of DNNs tends to be more inclined to the texture recognition of inputs, while in the deep layers it becomes very difficult to understand. 
%DenseNet~\cite{ref_a22}, Res152~\cite{ref_a21}, etc. Furthermore, in order to solve the mobile model storage and improve the model prediction speed, various lightweight models have also been proposed such as MobileNet~\cite{mob}. 

\begin{figure}
	\centering
	\includegraphics[height=5.3cm]{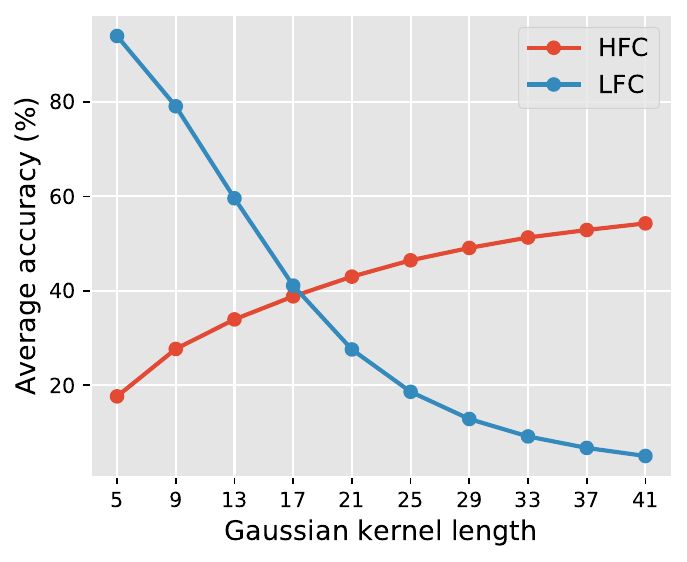}
	\caption{The average accuracy of HFC and LFC obtained by different Gaussian kernel. }
	\label{HFvsLF}
\end{figure}

To better understand the role of HFC and LFC for the classification results of DNNs, we split the information of raw images into these two pieces via Gaussian low-pass filter (defined in Eq.~\ref{g}). 
As illustrated in Fig.~\ref{HFvsLF}, when the kernel size is small, \textit{i.e.}, the cutoff frequency is high, the average accuracy of LFC on ten state-of-the-art models is close to 100\%. However, if we continue to increase the kernel size, the average accuracy of HFC begins to exceed LFC one. 
To our surprise, for several specific raw images, \textit{e.g.}, left image of Fig.~\ref{fig1}, the true label's confidence of HFC which is mostly black is even higher than the raw image.

To explain the above phenomenon, we turn to a perspective of feature space.
Interestingly, as shown in Fig.~\ref{featurespace}, 
most\footnote{see quantitative analysis in Appendix \textcolor{red}{A}}. feature maps in the shallow layers generally extract the edge and texture features (typical ones are highlighted by red boxes), \textit{i.e.}, HFC, and pay less attention to plain areas in images, \textit{i.e.}, LFC. Besides, recent feature-based attacks~\cite{zhou2018transferable,ganeshan2019fda,inkawhich2019feature} have shown that distortion in intermediate feature can lead to misclassification.
%\textbf{By contrast, due to the fact that distortion in intermediate feature space causes misclassification, \textit{e.g.} \cite{inkawhich2019feature}, deep layers are more responsible for fitting the ground-truth labels by selectively activating features of shallow layers. }
Therefore, low-level features, especially HFC-rich ones, are critical to classification. Put differently, 
\textit{if a perturbation can effectively manipulate the HFC of an image, then a model will extract completely different low-level features, which can lead to misclassification.}
% Inspired by recent intermediate feature-based attacks~\cite{zhou2018transferable,ganeshan2019fda,inkawhich2019feature}, we argue low-level features are critical to the classification. 
% % 	that the classification logic of DNNs can be split into two parts: the shallow layers and deep layers. 
% Interestingly, as shown in Fig.~\ref{featurespace}, 
% most\footnote{see quantitative analysis in Appendix ~\ref{S.B}.} feature maps in the shallow layers generally extract the edge and texture features (typical ones are highlighted by red boxes), \textit{i.e.}, HFC, and pay less attention to plain areas in images, \textit{i.e.}, LFC. %\textbf{By contrast, due to the fact that distortion in intermediate feature space causes misclassification, \textit{e.g.} \cite{inkawhich2019feature}, deep layers are more responsible for fitting the ground-truth labels by selectively activating features of shallow layers. }
% %  Therefore,HFC in shallow-layer features plays a crucial role in classification. Put differently, 
% \textit{Therefore, if a perturbation can effectively manipulate HFC of an image, totally different low-level features will be extracted and may lead to misclassification.}

\subsection{Effective Adversarial HFC}
\label{effectivehfc}
Before considering how to manipulate the images, we first investigate what kind of perturbations DNNs are more sensitive to since the performance of any other raw image's HFC is unsatisfactory (see the discussion in Appendix \textcolor{red}{H}).
% 	However, what kind of training-free noisy HFC can effectively fool DNNs is still unknown because the performance of any other raw image's HFC is unsatisfactory (see Sec.~5 in the Supplementary). %Note that the added HFC is equivalent to perturbing the (filtered) image with universal adversarial perturbation (UAP)~\cite{moosavi2017universal}.
A recent work~\cite{zhang2020understanding} has demonstrated that the effectiveness of adversarial perturbation lies in the fact that it contains irrelevant features and these features of perturbation dominate over the features in the raw image, thus leading to misclassification. Inspired by this finding, we intend to design adversarial HFC with strong irrelevant features, and we conjecture that the following properties are essential.

%However, what kind of training-free noisy HFC can effectively fool DNNs is still unknown because the performance of any other raw image's HFC is unsatisfactory (see Sec.~\textcolor{red}{5} in the appendix). Therefore, we empirically and experimentally analyze the properties required for effective adversarial HFC in this section. 

% However, what kind of training-free noisy HFC can effectively fool DNNs is still unknown because the performance of any other raw image's HFC is unsatisfactory (see Appendix ~\ref{raw}). %Note that the added HFC is equivalent to perturbing the (filtered) image with universal adversarial perturbation (UAP)~\cite{moosavi2017universal}.
% A recent work~\cite{zhang2020understanding} have demonstrated that the effectiveness of adversarial perturbation lies in the fact that it contains irrelevant features---features of perturbation dominate over features in the raw image, thus leading to misclassification. Inspired by their finding, we intend to design adversarial HFC with strong irrelevant features, and we conjecture that the following properties are essential.
        \textbf{Regionally Homogeneous.} Several recent works~\cite{li2020regional,pifgsm,dong2019evading} have demonstrated that adversarial perturbations with regionally homogeneous (or patch-wise) property can enhance the transferability of adversarial examples. 
		Considering that the raw image is a composite of homogeneous patterns, the reason might be attributed to that this perturbation is capable of forming irrelevant features recognizable by the DNNs.
		%The reason might be attributed to the fact that the homogeneous patterns tend to form features recognizable by the DNNs. %Inspired by the finding that adversarial examples contain non-robust features~\cite{bugs}, 

		\textbf{Repeating.}~\cite{ref_a33} observe that extra copies of the repeating element improve the confidence of DNNs. %Intuitively, repeating the content can strength the adversarial effect on the images.  %From the perspective of adversarial attack, it might help to enhance the attack ability of our adversarial examples. 
		From the perspective of strengthening the irrelevant features, we argue that repeating the content is beneficial.
		
		%\textbf{Dense.} 
		%Intuitively, dense adversarial HFC can affect more pixels of the original image and add more adversarial noises than sparse ones.
		%This is also witnessed by the green box in Fig.~\ref{featurespace} which can be viewed as a sparse adversarial HFC for ``dining table” but does not influence the top-1 label of classification.
		
		\textbf{Dense.} Analogous to the above \textit{repeating} property that performs global repeating, \textit{i.e.}, increases the amount of irrelevant features globally, 
		we can also perform local repeating to strengthen its adversarial effect further. For term distinction, we term this property \textit{dense}.

To verify the effect of the above properties, we conduct the ablation study in Sec.~\ref{ablationstudy}, and results support our conjecture. Besides, the analysis in Appendix \textcolor{red}{D} also shows that if the irrelevant features are similar to that of the target class, then our HIT has potential to become a targeted attack.

\subsection{Hybrid Image Transformation}
\label{uit}
%上面分析了需要什么性质，这边我们具体来介绍怎么做
%\textcolor{red}{Although original HFC play a key role in robustness, distorting LFC is also helpful to improve attack ability.} 

Motivated by the above discussion, we take the idea of hybrid image~\cite{art} to apply our no-box attacks. Formally, \cite{art} \textit{replaces} HFC of one image with HFC of another carefully picked image and craft hybrid images with two different interpretations: one that appears when the image is viewed up-close, and the other that appears from afar (see Fig. 5 of~\cite{art}). However, confusing human's vision system (without $\varepsilon$ constrain) cannot guarantee the misclassification of DNNs since adversarial examples are constrained by the maximum perturbation. 
Therefore, we propose a novel \textbf{Hybrid Image Transformation (HIT)} attack method which \textit{reduces}\footnote{Due to the $\varepsilon$ constraint, we cannot completely replace HFC with others} original HFC, and meanwhile, adds well-designed noisy ones to attack DNNs. Our method only needs three steps but can generate robust training-free adversarial perturbations in real time:

\begin{figure}[h]
	\centering
	\includegraphics[height=2.95cm]{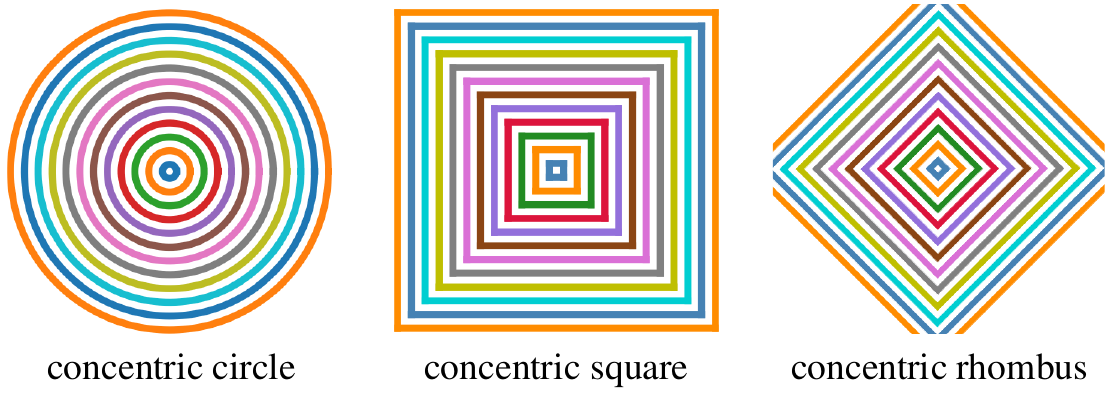}
	\caption{Three simple geometric patterns serve as proto-patterns.}
	\label{3tu}
\end{figure}
Firstly, we draw (\textit{i.e.}, training-free) the adversarial patch $\bm{x^p}$ to generate noisy HFC. 
% Unlike the traditional way that needs training, here we directly draw it. 
More specifically, as inspired by the observation in Sec.~\ref{effectivehfc}, we consider three simple \textbf{regionally homogeneous} proto-patterns (to avoid cherry-picking) as our basic adversarial patches: concentric circles, concentric squares, and concentric rhombus (shown in Fig.~\ref{3tu}). 
% Inspired by the observation in Sec.~\ref{effectivehfc},
The effect of concentric pattern is to make the resulting HFC \textbf{dense}. To satisfy the \textbf{repeating} property, we further resize and tile these adversarial patches (see the visualization in Appendix \textcolor{red}{E}).%Specifically, we first crop these proto-patterns to $300\times 300 \times 3$ adversarial patches, then resize them into different tile-sizes (\textit{e.g.}, $150\times 150 \times 3$) and tile them to $300\times 300 \times 3$, finally resize back to  $299\times 299\times 3$ to match the size of raw images. 

Secondly, we extract LFC of the raw image and HFC of the adversarial patch.
Note that several methods can be utilized to extract HFC and LFC of an image, \textit{e.g.}, Fourier transformation. In this paper, we use an approximated yet simple Gaussian low-pass filter $\bm{G}$ whose size is $(4k+1)\times (4k+1)$ to get LFC, which can be written as: 
\begin{equation}
\label{g}
	\bm{G_}{i,j}=\frac{1}{2\pi \sigma^2}e^{(-\frac{i^2+j^2}{2\sigma^2})},
\end{equation}
where $\sigma=k$ determines the \textit{width} of $\bm{G}$. Formally, the larger $\sigma$ is, the more HFC is filtered out. For simplicity, we are not going to introduce a new high-pass filter here and just get HFC by $\bm{G}$, \textit{i.e.}, obtaining HFC of the adversarial patch by subtracting its LFC.

Finally, we synthesize these two part components to generate our adversarial hybrid image $\bm{x^{adv}}$: %(shown in Fig.~\ref{show})
\begin{equation}
	\bm{x^{adv}}=\mathit{clip_{\bm{x},\varepsilon}}(\bm{x}*\bm{G}+\lambda \cdot (\bm{x^p} - \bm{x^p} * \bm{G})),
\end{equation}
where ``*” denotes convolution operation, $\lambda$ is a weight factor to balance LFC and HFC, and $clip_{x,\varepsilon}(\cdot)$ restricts resulting adversarial examples within $\varepsilon$-ball of the raw image in $l_\infty$ space. 
Please note that our HIT is different from patch attacks~\cite{advesarialpatch,eccvbias} which replace sub-regions of the image with adversarial patches.

As llustrated in Fig.~\ref{featurespace}(c), our HIT can effectively reduce relevant HFC and add many other irrelevant noisy ones, \textit{e.g.}, highlighted yellow boxes in (c). 
As a result, target model cannot extract correct features to make a reasonable prediction, thus leading to misclassification. Besides, our adversarial examples are less perceptible than those of our competitors.
% delete:(see Appendix \textcolor{red}{C}).

% \begin{figure*}[h] 
%   \begin{minipage}[t]{0.34\linewidth} 
%     \centering 
%     \includegraphics[height=4.4cm]{nips_img/SRvsR2.pdf}
%   \end{minipage}% 
%   \begin{minipage}[t]{0.32\linewidth} 
%     \centering 
%     \includegraphics[height=4.4cm]{nips_img/tile_results.pdf}
%   \end{minipage} 
% \begin{minipage}[t]{0.32\linewidth} 
%     \centering 
%     \includegraphics[height=4.4cm]{nips_img/density.pdf}
%   \end{minipage} 
% \centering
% \caption{The average attack success rates (\%) of ten models w.r.t the strength of semi-random noise $N_{sr}$ and random noise $N_{r}$ (\textbf{left}), tile-schemes (\textbf{middle}) and densities (\textbf{right}). ``blur” denotes using Gaussian kernel to smooth the image (constrained by maximum perturbation $\varepsilon$).}
% \label{3tufig}
% \end{figure*}

% \begin{figure*}[h]
% \centering
% \includegraphics[height=4.9cm]{nips_img/epsilon.pdf}
% \caption{The attack success rates (\%) of circle patches (\textbf{left}), square patches (\textbf{middle}) and rhombuses patches (\textbf{right}) w.r.t maximum perturbation $\varepsilon$.}
% \label{epsilon}
% \end{figure*} 
\begin{figure*}[t] 
      \begin{minipage}[t]{0.32\linewidth} 
        \centering 
        \includegraphics[height=5cm]{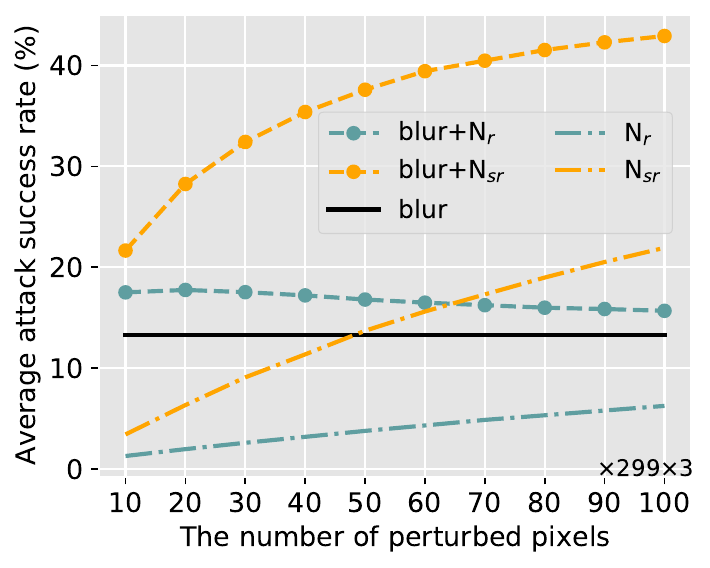}
      \end{minipage}% 
      \begin{minipage}[t]{0.3\linewidth} 
        \centering 
        \includegraphics[height=5cm]{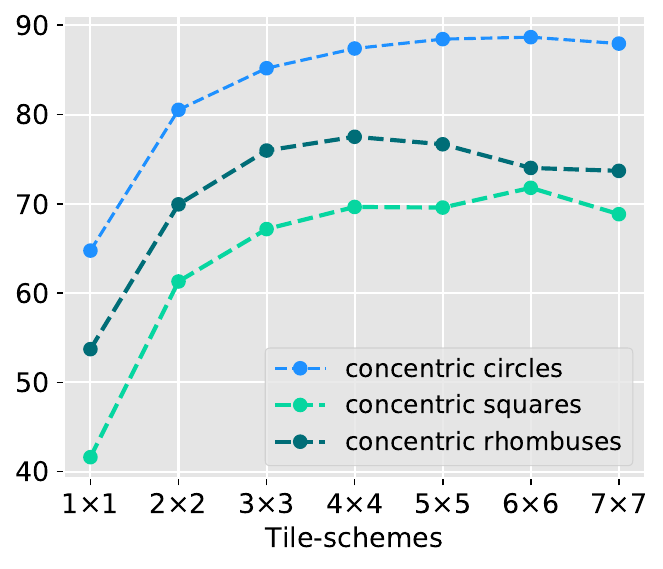}
      \end{minipage} 
    \begin{minipage}[t]{0.3\linewidth} 
        \centering 
        \includegraphics[height=5cm]{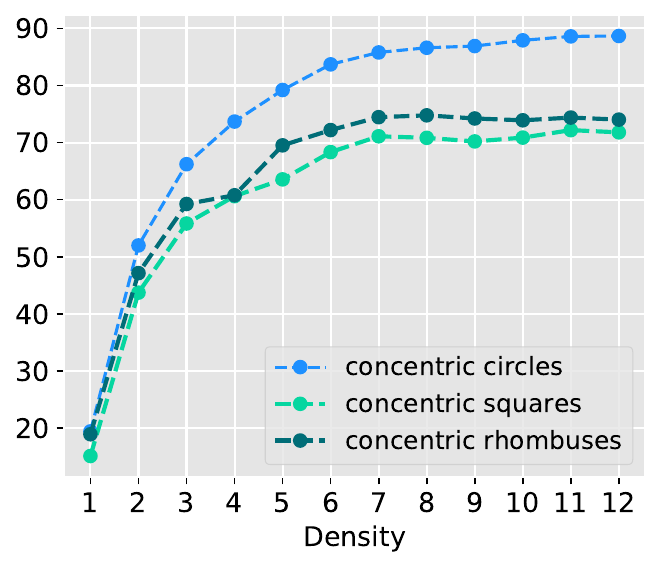}
      \end{minipage} 
    \centering
    \caption{The average attack success rates (\%) on normally trained models (for ImageNet) w.r.t. strength of semi-random noise $N_{sr}$ and random noise $N_{r}$ (\textbf{left}), tile-schemes (\textbf{middle}) and densities (\textbf{right}). ``blur” denotes using Gaussian kernel to smooth the image (constrained by $\varepsilon=16$).}
    \label{3tufig}
\end{figure*}

\section{Experiments}
\textbf{Target Model.} For ImageNet classification, we consider ten well-known normally trained classification models: VGG19~\cite{vgg}, Inception-v3 (Inc-v3)~\cite{inc-v3}, ResNet-152 (Res152)~\cite{resnet},  WideResNet-101 (WRN)~\cite{zagoruyko2016wide}, DenseNet-121 (Dense)~\cite{densenet}, SENet~\cite{hu2018squeeze}, PNASNet (PNA) \cite{pna}, ShuffleNet-v2 (Shuffle)~\cite{shuff}, SqueezeNet (Squeeze)~\cite{squee} and MobileNet-v2 (Mobile)~\cite{mob} as our target models. All the models are available in the Torchvision\footnote{\url{https://github.com/pytorch/vision/tree/master/torchvision/models}}, except for PNA and SENet which are obtained from Github\footnote{\url{https://github.com/Cadene/pretrained-models.pytorch}}. 
% Besides, we consider six defense models including three ensemble adversarial training models (EAT)~\cite{eat}: Inc-v3$_{ens3}$, Inc-v3$_{ens4}$ and IncRes-v2$_{ens}$ and three feature denoising models (FD)~\cite{featuredenoising}: ResNet152 Baseline (Res152$_B$), ResNet152 Denoise (Res152$_D$) and ResNeXt101 DenoiseAll (ResNeXt$_{DA}$).
For fine-grained classification, we use DCL framework~\cite{dcl} with three different backbones: ResNet50 (Res-50)~\cite{resnet}, SENet and SE-ResNet101 (SE-Res101)~\cite{resnet}.
We also consider a real-world recognition system---Google Cloud Vision API\footnote{\url{https://cloud.google.com/vision/docs/drag-and-drop}} in our experiment. Further, in Appendix \textcolor{red}{C}, we also consider other visual tasks, \textit{i.e.}, object detection and semantic segmentation.
% We also perform our attack on a real-world recognition system in Sec.~\ref{google_api}.

\textbf{Dataset.} For ImageNet classification, we choose 10,000 images from the ImageNet validation set~\cite{imagenet}. Each category contains about 10 images which are resized to $299\times 299 \times 3$ beforehand and are classified correctly by all normally trained networks. 
For fine-grained classification, we evaluate on the testing datasets ($448\times448\times3$) of CUB-200-2011~\cite{cub}, Stanford Cars~\cite{car} and FGVC Aircraft~\cite{air}.
% We also discuss our methods on other classification tasks in Appendix ~\ref{fine-grained_sec}.

\begin{table*}[t]

\centering
\resizebox{1\linewidth}{!}{
\begin{tabular}{ccp{1.1cm}<{\centering}p{1.1cm}<{\centering}p{1.1cm}<{\centering}p{1.1cm}<{\centering}p{1.1cm}<{\centering}p{1.1cm}<{\centering}p{1.1cm}<{\centering}p{1.1cm}<{\centering}p{1.1cm}<{\centering}p{1.1cm}<{\centering}p{1.1cm}<{\centering}}
	\toprule
	Model & Attack & VGG19 & Inc-v3 & Res152 & Dense & WRN & SENet & PNA  & Squeeze & Shuffle & Mobile & AVG. \\
	\midrule
	\multirow{3}[2]{*}{-} & 
	HIT (w/ Square) & 0.6215  & 0.7419  & 0.7638  & 0.8090  & 0.7599 & 0.5838  & 0.7437  & 0.6940  & 0.6704  & 0.4838  & 0.6872 \\
	& HIT (w/ Rhombus) & 0.6218  & 0.7458  & 0.7448  & 0.7853 & 0.7280  & 0.6258  & 0.7672  & 0.6738 & 0.6461 & 0.4005 & 0.6746 \\
	& HIT (w/ Circle) & \textbf{0.5472}  & \textbf{0.6685}  & \textbf{0.7306}  & \textbf{0.7779}  & \textbf{0.7223}  & \textbf{0.5613}  & \textbf{0.6747}  & \textbf{0.6643} & \textbf{0.6062}  & \textbf{0.3617}  & \textbf{0.6314}\\
	\bottomrule
\end{tabular}}%
\caption{The cosine similarity comparison on low-level features of models for different patterns. The maximum perturbation $\varepsilon=16$. The bold value denotes the lowest similarity.}
\label{cosine}
\end{table*}%

\textbf{Parameters.} In our experiments, we use $l_\infty$-norm to measure perceptibility of adversarial noises and maximum $\varepsilon$ is set according to the default settings of comparison methods. 
%unless specified, the maximum perturbation $\varepsilon$ is set to 16. 
For the \textit{no-box threat model}, maximum perturbation $\varepsilon$ is set to 25.5 ($0.1\times255$) and parameters of~\cite{li2020practical} follow its default setting.
For our HIT, size of Gaussian kernel $\bm{G}$ is $17\times 17$ (\textit{i.e.} $k=4$), weight factor $\lambda$ is set to 1.0 (discussion about $\lambda$ is shown in Appendix \textcolor{red}{B}), and density of proto-pattern is set to 12. For tile-size, unless specified, we set to $50\times 50$, \textit{i.e.}, tile-scheme is $6\times6$.
For the \textit{traditional black-box threat model}, iteration $T$ is set to 10, maximum perturbation $\varepsilon=16$ and step size $\alpha$ is 1.6. In addition, we adopt the default decay factor $\mu=1.0$ for MI-FGSM~\cite{dong2018boosting}. we set the transformation probability to 0.7 for DI$^2$-FGSM~\cite{xie2019improving}. Length of Gaussian kernel for TI-FGSM~\cite{dong2019evading} is 15.
For PI-FGSM~\cite{pifgsm}, length of project kernel is 3, amplification factor $\beta$ and project factor $\gamma$ are 10.0 and 16.0, respectively. For S$^2$I-FGSM~\cite{long2023frequency}, the number of spectrum transformations $N = 20$.
%Different from PI-FGSM, $\beta$ and $\gamma$ for PI-MI-DI$^2$-FGSM and PI-TI-DI$^2$-FGSM~\cite{pifgsm} is 2.5 and 2.0, respectively.
For the \textit{practical black-box threat model}, maximum perturbation $\varepsilon$ is set to 10. For CDA~\cite{cda} and BIA~\cite{Zhang2022BIA}, resulting generators are trained on 1.2 million ImageNet training dataset (using Adam optimizer~\cite{kingma2014adam} with a learning rate of 2e-4) for one epoch with batch size 16.
For DR~\cite{dr} and SSP~\cite{ssp}, we set step size $\alpha=4$, iterations $T=100$. For DR, SSP and BIA which perturb low-level features, we attack the output of $Maxpool.3$ for VGG-19 and the output of $Conv3\_8$ for Res-152.

\subsection{Ablation Study}
\label{ablationstudy}
% In this section, we conduct a series of ablation study for our HIT. Specifically, we demonstrate the effectiveness of regionally homogeneous pattern, repeating pattern and dense pattern.
% Besides, we also give an insight into our proto-pattern.
% For the result of HIT without reducing HFC beforehand is shown in Appendix I.
\begin{table*}[h]

\centering
\resizebox{1\linewidth}{!}{
	\begin{tabular}{cp{1.2cm}<{\centering}p{1.2cm}<{\centering}p{1.2cm}<{\centering}p{1.2cm}<{\centering}p{1.2cm}<{\centering}p{1.2cm}<{\centering}p{1.2cm}<{\centering}p{1.2cm}<{\centering}p{1.2cm}<{\centering}p{1.2cm}<{\centering}p{1.2cm}<{\centering}}
		\toprule
		Attack & VGG19 & Inc-v3 & Res152 & DenseNet & WRN & SENet & PNA   & Shuffle & Squeeze & Mobile & AVG. \\
		\midrule
		Naïve$^\ddagger$ w/o Sup. & 54.08  & 36.06  & 39.36  & 43.52  & 41.20  & 34.46  & 26.86  & -     & -     & 62.24  & 42.22  \\
		Jigsaw w/o Sup. & 68.46  & 49.72  & 53.76  & 57.62  & 48.76  & 40.94  & 37.68  & -     & -     & 74.76  & 53.96  \\
		Rotation w/o Sup. & 68.86  & 51.86  & 52.60  & 58.74  & 49.28  & 41.80  & 40.06  & -     & -     & 74.00  & 54.65  \\
		\midrule
		Naïve$^\dagger$ w/ Sup. & 23.80  & 19.14  & 16.24  & 21.06  & 15.84  & 13.00  & 13.04  & -     & -     & 27.56  & 18.71  \\
		Prototypical w/ Sup. & 80.22  & 63.54  & 62.08  & 70.84  & 62.72  & 55.44  & 51.42  & -     & -     & 82.22  & 66.06  \\
		Prototypical$^*$ w/ Sup. & 81.26  & 66.32  & 65.28  & 73.94  & 66.86  & 57.64  & 54.98  & -     & -     & 83.66  & 68.74  \\
		\midrule
		Beyonder w/ Sup. & 75.04  & 48.88  & 69.40  & 72.88  & 66.06  & 56.22  & 48.20  & -     & -     & 72.98  & 63.71  \\
		\midrule
		HIT (Ours) & \textbf{99.30} & \textbf{98.92} & \textbf{98.55} & \textbf{98.69} & \textbf{96.53} & \textbf{93.61} & \textbf{97.02} & \textbf{99.56} & \textbf{99.63} & \textbf{99.52} & \textbf{98.13} \\
		\bottomrule
\end{tabular}}%
\caption{The comparison of attack success rates (\%) between state-of-the-art no-box attacks and ours with the maximum perturbation $\varepsilon=25.5$ (Sup. means supervised mechanism). The bold value denotes the best transferability.}
\label{no-box}%
\end{table*}%
\begin{table*}[h]
\centering
\resizebox{1\linewidth}{!}{
\begin{tabular}{ccp{1.1cm}<{\centering}p{1.1cm}<{\centering}p{1.1cm}<{\centering}p{1.1cm}<{\centering}p{1.1cm}<{\centering}p{1.1cm}<{\centering}p{1.1cm}<{\centering}p{1.1cm}<{\centering}p{1.1cm}<{\centering}p{1.1cm}<{\centering}p{1.1cm}<{\centering}}
	\toprule
	Model & Attack & VGG19 & Inc-v3 & Res152 & Dense & WRN & SENet & PNA  & Squeeze & Shuffle & Mobile & AVG. \\
	\midrule
	\multirow{4}[2]{*}{VGG19} 
	& MI-FGSM& 99.96*  & 23.92  & 30.82  & 54.54  & 28.94  & 36.81  & 32.86  & 69.87  & 47.15  & 58.50  & 48.34  \\
	& DI$^2$-FGSM & 99.96*  & 14.29  & 27.80  & 47.95  & 24.53  & 32.93  & 23.19  & 40.08  & 27.90  & 53.01  & 39.16  \\
	& PI-FGSM & 99.95*  & 36.22  & 36.46  & 55.39  & 39.40  & 35.28  & 50.84  & 81.24  & 60.26  & 69.89  & 56.49  \\
% 	& PI-MI-DI$^2$-FGSM~\cite{pifgsm} & 99.96*  & 47.23  & 59.01  & 80.39  & 57.05  & 65.17  & 55.25  & 83.20  & 63.92  & 83.49  & 66.08  \\
	& S$^2$I-FGSM &99.99* &	40.12 &	47.68 &	68.93 &	41.95 &	38.41 &	56.30 &	67.22	 &51.88 &	76.85 &	58.93\\
	\midrule
	\multirow{4}[2]{*}{Inc-v3} 
	& MI-FGSM& 42.58  & 99.92*  & 33.95  & 42.30  & 33.43  & 27.57  & 41.93  & 68.34  & 51.22  & 53.05  & 49.43  \\
	& DI$^2$-FGSM & 33.91  & 99.33*  & 24.34  & 32.69  & 21.83  & 19.18  & 30.39  & 35.90  & 29.35  & 34.87  & 36.18  \\
	& PI-FGSM & 51.77  & 99.91*  & 35.56  & 50.44  & 38.67  & 31.78  & 52.07  & 78.34  & 58.96  & 62.53  & 56.00  \\
% 	& PI-MI-DI$^2$-FGSM~\cite{pifgsm} & 68.27  & 99.76*  & 56.64  & 70.09  & 57.53  & 51.52  & 61.86  & 80.76  & 67.26  & 74.01  & 65.33  \\
	& S$^2$I-FGSM & 51.47 &	99.84* &	43.47 &	52.28 &	38.16 &	29.05 &	54.55 &	49.24 &	44.86 &	55.89 &	51.88\\
	\midrule
	\multirow{4}[2]{*}{Res152} 
	& MI-FGSM& 63.75  & 41.71  & 99.98*  & 72.93  & 85.27  & 49.99  & 46.56  & 75.86  & 65.63  & 72.40  & 67.41 \\
	& DI$^2$-FGSM & 76.90  & 41.22  & 99.95*  & 82.16  & 88.35  & 60.23  & 44.73  & 58.88  & 60.24  & 76.22  & 68.89 \\
	& PI-FGSM & 64.88  & 48.16  & 99.98*  & 68.92  & 79.49  & 45.23  & 61.37  & 82.94  & 71.18  & 76.32  & 69.85 \\
% 	& PI-MI-DI$^2$-FGSM~\cite{pifgsm} & 92.71  & 77.90  & 99.99*  & \textbf{96.04}  & \textbf{97.80}  & \textbf{86.09}  & 78.00  & 90.75  & 86.76  & 93.51  & \textbf{88.84}  \\
	& S$^2$I-FGSM &88.67&	69.9	&99.99*	&92.02	&\textbf{96.41}&	62.98&	75.95&	74.77	&79.38	&91.66	&83.17 \\
	\midrule
	\multirow{4}[2]{*}{Dense} 
	& MI-FGSM& 76.89  & 46.00  & 69.86  & 99.98*  & 67.76  & 53.05  & 48.69  & 78.55  & 69.63  & 78.42  & 68.88 \\
	& DI$^2$-FGSM & 81.14  & 35.96  & 69.39  & 99.98*  & 64.64  & 48.53  & 40.03  & 60.50  & 55.68  & 73.03  & 62.89  \\
	& PI-FGSM & 74.55  & 52.09  & 61.22  & 99.98*  & 63.12  & 49.84  & 60.09  & 85.79  & 74.74  & 82.37  & 70.38  \\
% 	& PI-MI-DI$^2$-FGSM~\cite{pifgsm} & 94.46  & 74.76  & \textbf{90.88}  & 99.99*  & 89.42  & 80.99  & 73.86  & 91.00  & 85.34  & 93.93  & 86.07  \\
    & S$^2$I-FGSM & \textbf{95.28} &74.79 & \textbf{91.46} &99.99* &86.83 &68.43 & 74.39 & 80.65 & 81.72 & 92.57 & 84.61\\
	\midrule
	\multirow{1}[1]{*}{-}
	& HIT (Ours) & 94.75  & \textbf{90.37}  & 87.62  & \textbf{88.81}  & 79.26  & \textbf{70.31}  & \textbf{82.12}  & \textbf{98.31}  & \textbf{97.34}  & \textbf{97.81}  & \textbf{88.67}  \\
	\bottomrule
\end{tabular}}%
\caption{The comparison of attack success rates (\%) on normally trained models between black-box attacks (``*” denotes white-box attack) and our no-box attacks with the maximum perturbation $\varepsilon=16$. The bold value denotes the best transferability.}
\label{nt}%
\end{table*}%

\begin{table*}[h]
\centering
\resizebox{1\linewidth}{!}{
\begin{tabular}{ccp{1.3cm}<{\centering}p{1.3cm}<{\centering}cp{1.3cm}<{\centering}p{1.3cm}<{\centering}cp{1.3cm}<{\centering}p{1.3cm}<{\centering}cc}
\toprule
\multirow{2}{*}{Model} & \multirow{2}{*}{Attacks} & \multicolumn{3}{c}{CUB-200-2011} & \multicolumn{3}{c}{Stanford Cars} & \multicolumn{3}{c}{FGVC Aircraft} &  \multirow{2}{*}{AVG.} \\
 &  & Res50 & SENet & SE-Res101 & Res50 & SENet & SE-Res101 & Res50 & SENet & SE-Res101  \\
 \midrule
\multirow{4}{*}{VGG19} & DR & 7.46 & 6.04 & 5.33 & 3.55 & 3.35 & 1.96 & 8.46 & 6.65 & 7.98 & 5.64 \\
 & SSP & 27.95 & 32.80 & 18.61 & 34.39 & 18.31 & 9.90 & 36.27 & 24.50 & 27.23 & 25.55 \\
 & CDA & 31.91 & 29.64 & 20.86 & 37.97 & 24.27 & 13.20 & 35.75 & 43.27 & 32.25 & 29.90 \\
 & BIA & 44.02 & 39.72 & 34.76 & 29.69 & 22.40 & 19.24 & 44.76 & 41.31 & 43.65 & 35.50 \\
 \midrule
\multirow{4}{*}{Res152} & DR & 11.18 & 5.46 & 6.89 & 6.77 & 3.48 & 2.47 & 15.55 & 10.65 & 12.86 & 8.37 \\
 & SSP & 45.46 & 23.78 & 21.56 & 43.52 & 15.28 & 10.24 & 36.73 & 20.33 & 24.55 & 26.83 \\
 & CDA & 48.31 & 38.15 & \textbf{38.93} & 38.82 & 32.42 & 21.43 & 29.66 & 49.24 & 35.68 & 36.96 \\
 & BIA & \textbf{50.14} & \textbf{42.98} & 35.84 & \textbf{66.45} & \textbf{42.03} & 27.71 & \textbf{58.27} & \textbf{64.75} & \textbf{44.30} & \textbf{48.05} \\
 \midrule
\multirow{1}{*}{-} 
 & HIT (Ours) & 44.77 & 24.53 & 32.09 & 64.95 & 26.17 & \textbf{38.00} & 41.67 & 35.48 & 41.99 & 38.85 \\
 \bottomrule
\end{tabular}}
\caption{The comparison of attack success rates (\%) on fine-grained classification tasks between practical black-box attacks (the leftmost column is the ImageNet pre-trained substitute model) and our no-box attacks with the maximum perturbation $\varepsilon=10$. The bold value denotes the best transferability.}
\label{fine-grained}
\end{table*}
\subsubsection{The Effect of Regionally Homogeneous Pattern}
\label{random}
To the best of our knowledge, regionally homogeneous perturbations~\cite{dong2019evading,pifgsm,pifgsm++,li2020regional} are mostly based on the gradient to craft, thereby training is necessary. However, whether arbitrary noise can benefit from the homogeneous property remains unclear. Thus, we compare random noises with semi-random ones to check it:

\textbf{Random noise $\bm{N_r}$:} For a given random location pair set $L$, we call $\bm{N_r}\in \mathbb{R}^{H\times W \times C}$ as random noise if it satisfies the following definition:
	\begin{equation}
		\bm{N_r}[i,j,c] =\begin{cases}
			\varepsilon \cdot \mathit{random}(-1, 1),&  (i,j,c)\in L \\
			0 ,& else
		\end{cases}
	\end{equation}
	
\textbf{Semi-random noise $\bm{N_{sr}}$:} Different from the random noise, semi-random noise has some regularity. Let $S$ denote a semi-random location pair set (here we take $H$-dimension random noise as an example), $\bm{N_{sr}}$ can be written as:
	\begin{equation}
		\bm{N_{sr}}[i,:,:] =\begin{cases}
			\varepsilon \cdot \mathit{random}(-1, 1) ,&  i\in S \\
			0 ,& else
		\end{cases}
		\label{semi-random}
	\end{equation}
	
where \textit{random}(-1, 1) returns 1 or -1 randomly. As depicted in Fig.~\ref{3tufig}, success rates of $N_{sr}$ are consistently higher than those of $N_r$. As number of perturbed pixels increases, the margin between them also increases.
This demonstrates that training-free noise can also benefit from \textbf{regionally homogeneous} property.
To exploit this conclusion further, in Fig.~\ref{3tu}, we extend semi-random noise to other more complex ``continuous” patterns, \textit{e.g.}, circle.

\subsubsection{The Effect of Repeating Pattern}
\label{repeat}
In this section, we show experimental results of our proposed HIT w.r.t. tile-sizes. Here we consider seven different tile-schemes including $1\!\times\!1$, $2\!\times\!2$, $3\!\times\!3$, $4\!\times\!4$, $5\!\times\!5$, $6\!\times\!6$ and $7\!\times\!7$, and tile-sizes thereby are $300\!\times\!300$, $150\!\times\!150$, $100\!\times\!100$, $75\!\times\!75$, $60\!\times \!60$, $50\!\times\!50$, $42\!\times\!42$, respectively. Please note we will resize back to $299\!\times\!299\!\times3$ to match the size of raw images. The visualizations of these patches can be found in Appendix \textcolor{red}{E}.

In Fig.~\ref{3tufig}, we report average attack success rates on ten models. The success rates increase very quickly at first and then keep stable after $4\!\times\!4$ tile-scheme. This result demonstrates that attack ability of training-free perturbations can benefit from \textbf{repeating} property. If we continue to increase tile-size, attack success rates may go down. The main reason might be that distortion caused by the resizing operation---It indirectly blurs tiled adversarial patches, thus reducing available HFC.

\subsubsection{The Effect of Dense Pattern}
\label{dense}
To validate the effect of dense pattern, we analyze the average attack success rates w.r.t. densities. Since trends of different patterns are similar, we only discuss the results of circle patch whose tile-scheme is $6\times6$. 
Here we control the density from 1 to 12. For example, ``2” denotes only two circles in the proto-pattern, and more visualizations can be found in Appendix \textcolor{red}{E}.  

As shown in Fig.~\ref{3tufig}, the success rates increase rapidly at the beginning, then remain stable after the density exceeds 8, and reach the peak at 12. This experiment demonstrates the effectiveness of \textbf{dense} pattern. Therefore, we set the default density of each proto-pattern to 12 in our paper.

\subsubsection{The Size of Perturbation}
\label{sizeofp}
\begin{figure*}[h]
\centering
\includegraphics[height=5.3cm]{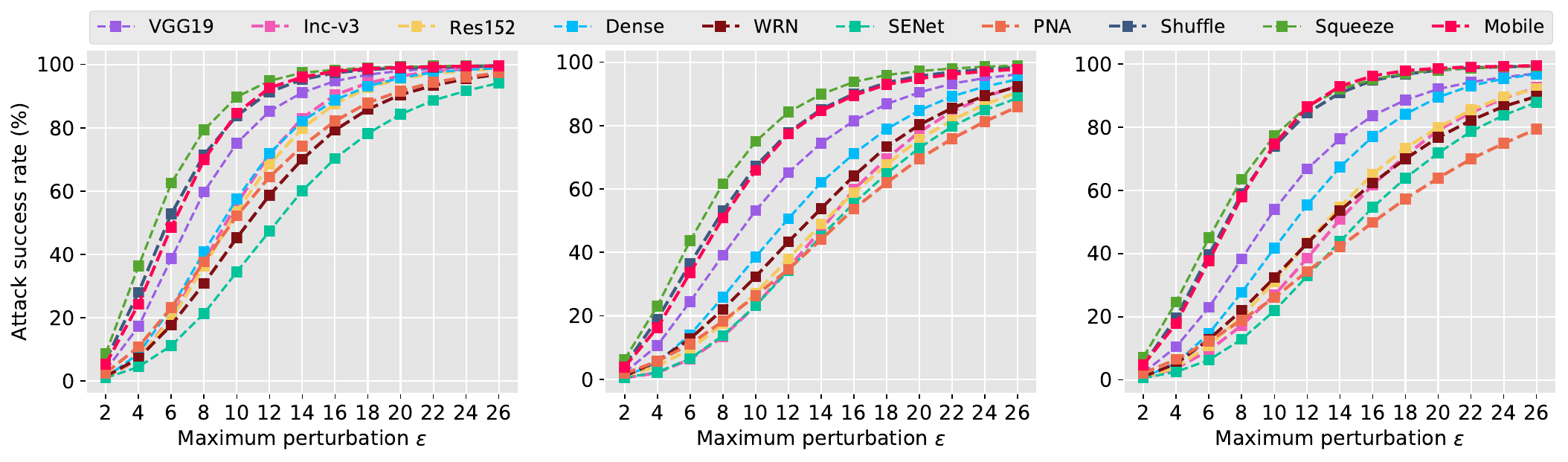}
\caption{The attack success rates (\%) of circle patches (\textbf{left}), square patches (\textbf{middle}) and rhombuses patches (\textbf{right}) on normally trained models w.r.t. maximum perturbation $\varepsilon$.}
\label{epsilon}
\end{figure*}

In this section, we study the influence of the maximum perturbation $\varepsilon$ on the performance of our HIT. 
The result of Fig.~\ref{epsilon} depicts growth trends of each model under different adversarial patches. 
No matter what the adversarial patch is, the performance proliferates at first, then remains stable after $\varepsilon$ exceeds 16 for most models.

\subsubsection{The Analysis of Proto-patterns}

As shown in Fig.~\ref{3tufig}, the Circle usually performs best. To help understand this phenomenon, we provide an insight from a perspective of intermediate feature response. Without loss of generality, we set the layer index to ``depth of each model" / 2 and report the average cosine similarity of the features between 10,000 raw images and corresponding adversarial examples. The result from Tab.~\ref{cosine} shows that the Circle consistently leads to lower cosine similarity than other patterns. Consequently, the features that feed to deep layer are more featureless, thus leading to misclassification more easily. As for why the Circle is more effective in decreasing cosine similarity is beyond the scope of this paper, but may be an interesting direction for future research.

For the sake of simplicity, we only report the results of HIT based on the Circle in the following.

\subsection{HIT vs. No-box Attacks}
In this section, we compare performance of our no-box HIT with state-of-the-art no-box attacks~\cite{li2020practical}. Note that~\cite{li2020practical} need to pay 15,000 iterations at most to train a substitute model, and then runs extra 200 iterations baseline attacks and 100 iterations ILA~\cite{huang2019enhancing}, which is highly time-consuming. 
Different from~\cite{li2020practical}, our HIT is training-free, which does not require any auxiliary images to train a substitute model or run extra iterations to craft adversarial examples, thus being real-time.

The experimental results are reported in Tab.~\ref{no-box}. A first glance shows that HIT outperforms the existing state-of-the-art no-box attacks by a significantly large margin. Notably, HIT can achieve an average success rate of \textbf{98.13\%}. By contrast, the best performance of~\cite{li2020practical}, \textit{i.e.}, Prototypical$^*$ w/ Sup, is only 68.74\% on average. This convincingly demonstrates the effect of our design.

\subsection{HIT vs. Traditional Black-box Attacks}
In this section, we compare our no-box HIT with traditional transfer-based attacks.
For MI-FGSM, DI$^2$-FGSM, PI-FGSM and S$^2$I-FGSM, we utilize VGG19, Inc-v3, Res152 and Dense to iteratively craft adversarial examples and use them to attack target models. As for our HIT, it does not need any substitute model or training process.
The results are summarized in Tab.~\ref{nt}, where the models in the leftmost column are substitute models, and the bottom block shows results of our HIT. 

Notably, our HIT is even on par with the most effective S$^2$I-FGSM in Tab.~\ref{nt}.
% As demonstrated in Tab.~\ref{nt}, our HIT is even on par with most effective PI-MI-DI$^2$-FGSM. 
Specifically, on average, the best performance of S$^2$I-FGSM is 84.61\%, while our HIT can get up to 88.67\%. Intuitively, the transferability of adversarial examples largely depends on the substitute model. For example, when adversarial examples are crafted via VGG19, performance of S$^2$I-FGSM is limited and our HIT can remarkably outperform it by \textbf{29.74\%} on average. Besides, when the target model is a lightweight model, \textit{e.g.}, Shuffle, our method can consistently outperform these transfer-based attacks by a large margin. Furthermore, results in Appendix \textcolor{red}{F}\&\textcolor{red}{G} also demonstrate our HIT can evade defenses.

\subsection{HIT vs. Practical Black-box Attacks}
\label{fine-grained_sec}
To highlight the practical property of our HIT, we apply HIT to attack other classification tasks. Specifically, we consider three well-known fine-grained datasets: CUB-200-2011~\cite{cub}, Stanford Cars~\cite{car} and FGVC Aircraft~\cite{air}, and the victim model is trained via DCL~\cite{dcl}. Due to resolution of inputs is $448\times448$, we set tile-size equal to $448\times448$ divided by tile-scheme.
Please note other parameters for HIT still follow the default setting.

As demonstrated in Tab.~\ref{fine-grained}, our proposed no-box attack is also on par with state-of-the-art practical black-box methods. For example, HIT results in 38.85\% of images being misclassified by these models, while average attack success rates (substitute model is VGG19) obtained by DR, SSP and CDA are 5.64\%, 25.55\% and 29.90\%, respectively. Even compared to 
stronger BIA, our proposed method is not inferior in some cases. Remarkably, \textbf{38.00\%} adversarial examples crafted by our HIT can fool DCL (backbone: SE-Res101) trained on the Stanford Cars, while BIA only gets a 27.71\% success rate when the substitute model is Res152.

\begin{figure}[h]
% \vspace{-0.5cm}
\centering
\includegraphics[height=5cm]{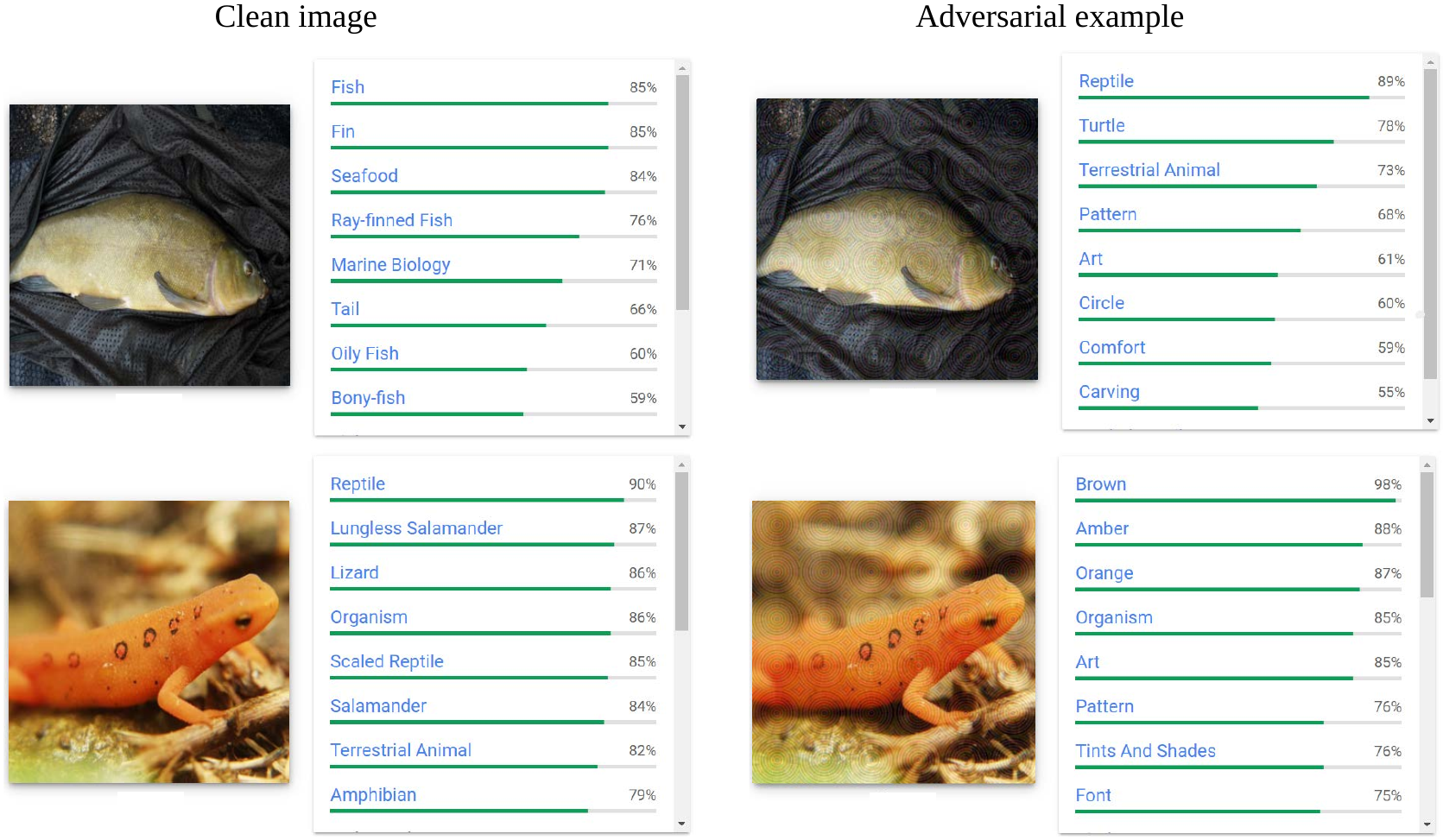}
\caption{The results for attacking Google Cloud Vision API. The maximum perturbation $\varepsilon$ is 16.}
\label{adv_real}
\end{figure}
\subsection{Attack Real-world Recognition System}
\label{google_api}

To further demonstrate the practical property of our HIT, we attack a real-world recognition system, \textit{i.e.}, Google Cloud Vision API. Different from existing works~\cite{chen2017zoo,brendel2017decision} which need a large number of queries for optimization, we can directly apply HIT with the default setting to craft adversarial example.
In this experiment, we randomly sample 200 raw images that are correctly classified by the API to 
evaluate our HIT($\varepsilon=16$). Notably, our attack success rate can reach \textbf{72.5\%}. Besides, we observe that our no-box HIT  can effectively change top-k labels. As illustrated in Fig.~\ref{adv_real}, the top-5 label of the Fish (top row) is {``Fish", ``Fin", ``Seafood", ``Ray-finned fish" and ``Marine biology"}, while our adversarial example is {``Reptile",``Turtle", ``Terrestrial Animal", ``Pattern" and ``Art"}.
This result also demonstrates that our HIT can effectively disrupt category-dependent features.
\begin{figure*}[h]
\centering
\includegraphics[height=12cm]{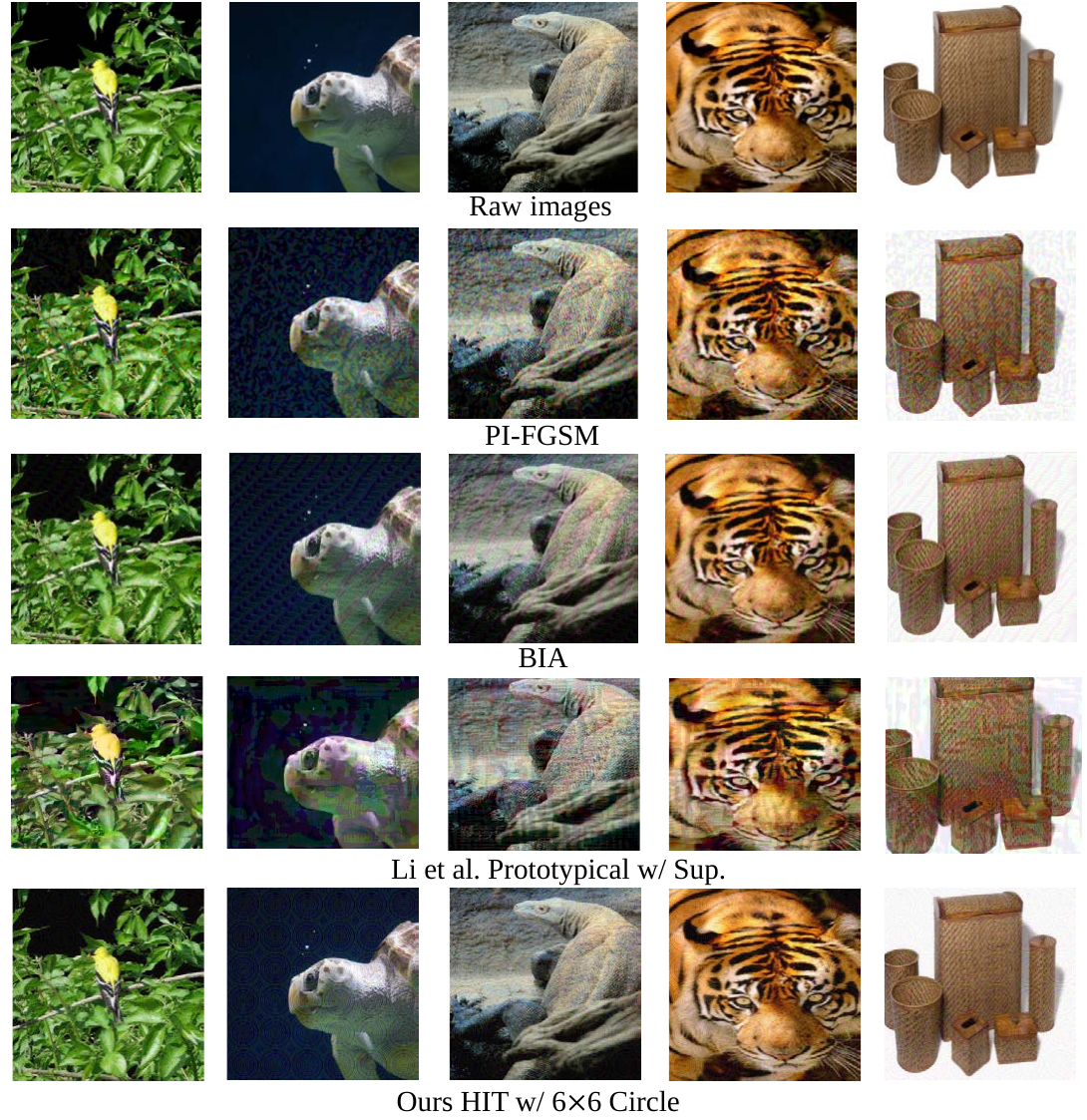}
\caption{Qualitative comparison for adversarial examples ($\varepsilon=16$) crafted by different methods.}
\label{adv_cmp}
\end{figure*}
\begin{equation}
\bm{x^H_r}=\bm{x_r}-\bm{x_r}*\bm{G}.
\end{equation}

\section{Qualitative Comparison for Adversarial Examples}
\label{qualitative}

To better reflect advantages of our approach, in this section, we compare visual quality of the generated adversarial examples. Specifically, we consider black-box PI-FGSM~\cite{pifgsm}, practical black-box BIA~\cite{Zhang2022BIA} and no-box attack~\cite{li2020practical} as our competitors. As depicted in Fig.~\ref{adv_cmp}, PI-FGSM~\cite{pifgsm}, BIA~\cite{Zhang2022BIA} and no-box attack~\cite{li2020practical} usually cause more perceptible distortions. In contrast, the adversarial perturbation crafted by our HIT is less perceptible.

\section{Conclusion}
\label{conclusion}
In this paper, we rethink the classification logic of deep neural networks with respect to adversarial examples. We observe that HFC is dominant in low-level features and plays a crucial role in classification. Besides, we demonstrate that DNNs are vulnerable to training-free perturbations with regionally homogeneous, repeating, dense properties through empirically and experimentally analysis. Motivated by these observations, we propose a novel Hybrid Image Transformation (HIT) attack method which combines LFC of raw images with HFC of our well-designed adversarial patches to destroy useful features and add strong irrelevant noisy ones. 
Remarkably, our method outperforms existing no-box attacks by a significantly large margin and is even on par with transfer-based black-box attacks. 

\bibliographystyle{IEEEtran}
\bibliography{main}

\begin{thebibliography}{10}
\providecommand{\url}[1]{#1}
\csname url@samestyle\endcsname
\providecommand{\newblock}{\relax}
\providecommand{\bibinfo}[2]{#2}
\providecommand{\BIBentrySTDinterwordspacing}{\spaceskip=0pt\relax}
\providecommand{\BIBentryALTinterwordstretchfactor}{4}
\providecommand{\BIBentryALTinterwordspacing}{\spaceskip=\fontdimen2\font plus
\BIBentryALTinterwordstretchfactor\fontdimen3\font minus \fontdimen4\font\relax}
\providecommand{\BIBforeignlanguage}[2]{{%
\expandafter\ifx\csname l@#1\endcsname\relax
\typeout{** WARNING: IEEEtran.bst: No hyphenation pattern has been}%
\typeout{** loaded for the language `#1'. Using the pattern for}%
\typeout{** the default language instead.}%
\else
\language=\csname l@#1\endcsname
\fi
#2}}
\providecommand{\BIBdecl}{\relax}
\BIBdecl

\bibitem{vgg}
K.~Simonyan and A.~Zisserman, ``Very deep convolutional networks for large-scale image recognition,'' in \emph{ICLR}, Y.~Bengio and Y.~LeCun, Eds., 2015.

\bibitem{inc-v3}
C.~Szegedy, V.~Vanhoucke, S.~Ioffe, J.~Shlens, and Z.~Wojna, ``Rethinking the inception architecture for computer vision,'' in \emph{CVPR}, 2016.

\bibitem{inc-v4}
C.~Szegedy, S.~Ioffe, V.~Vanhoucke, and A.~A. Alemi, ``Inception-v4, inception-resnet and the impact of residual connections on learning,'' in \emph{AAAI}, 2017.

\bibitem{resnet}
K.~He, X.~Zhang, S.~Ren, and J.~Sun, ``Deep residual learning for image recognition,'' in \emph{CVPR}, 2016.

\bibitem{densenet}
G.~Huang, Z.~Liu, L.~van~der Maaten, and K.~Q. Weinberger, ``Densely connected convolutional networks,'' in \emph{CVPR}, 2017.

\bibitem{wang}
X.~Wang, Y.~Guo, J.~Song, L.~Gao, and H.~T. Shen, ``Amanet: Adaptive multi-path aggregation for learning human 2d-3d correspondences,'' \emph{IEEE Transactions on Multimedia}, vol.~25, pp. 979--992, 2023.

\bibitem{10704987}
X.~Wang, X.~Chen, L.~Gao, J.~Song, and H.~T. Shen, ``Cpi-parser: Integrating causal properties into multiple human parsing,'' \emph{IEEE Transactions on Image Processing}, vol.~33, pp. 5771--5782, 2024.

\bibitem{TMM4}
J.~Wang, J.~Zhao, Q.~Yin, X.~Luo, Y.~Zheng, Y.-Q. Shi, and S.~K. Jha, ``Smsnet: A new deep convolutional neural network model for adversarial example detection,'' \emph{IEEE Transactions on Multimedia}, vol.~24, pp. 230--244, 2022.

\bibitem{TMM5}
R.~Ran, J.~Wei, C.~Zhang, G.~Wang, Y.~Yang, and H.~T. Shen, ``Adaptive multi-scale degradation-based attack for boosting the adversarial transferability,'' \emph{IEEE Transactions on Multimedia}, pp. 1--12, 2024.

\bibitem{TMM6}
L.~Gao, Z.~Huang, J.~Song, Y.~Yang, and H.~T. Shen, ``Push \& pull: Transferable adversarial examples with attentive attack,'' \emph{IEEE Transactions on Multimedia}, vol.~24, pp. 2329--2338, 2022.

\bibitem{TMM7}
X.~Wei and S.~Zhao, ``Boosting adversarial transferability with learnable patch-wise masks,'' \emph{IEEE Transactions on Multimedia}, vol.~26, pp. 3778--3787, 2024.

\bibitem{TMM8}
X.~Wang, H.~Chen, P.~Sun, J.~Li, A.~Zhang, W.~Liu, and N.~Jiang, ``Advst: Generating unrestricted adversarial images via style transfer,'' \emph{IEEE Transactions on Multimedia}, vol.~26, pp. 4846--4858, 2024.

\bibitem{gaker}
Y.~Sun, S.~Yuan, X.~Wang, L.~Gao, and J.~Song, ``Any target can be offense: Adversarial example generation via generalized latent infection,'' in \emph{ECCV}, 2024.

\bibitem{szegedy2013intriguing}
C.~Szegedy, W.~Zaremba, I.~Sutskever, J.~Bruna, D.~Erhan, I.~J. Goodfellow, and R.~Fergus, ``Intriguing properties of neural networks,'' in \emph{ICLR}, 2014.

\bibitem{goodfellow2014explaining}
I.~J. Goodfellow, J.~Shlens, C.~Szegedy \emph{et~al.}, ``Explaining and harnessing adversarial examples,'' in \emph{ICLR}, 2015.

\bibitem{madry2017towards}
A.~Madry, A.~Makelov, L.~Schmidt, D.~Tsipras, and A.~Vladu, ``Towards deep learning models resistant to adversarial attacks,'' in \emph{ICLR}, 2018.

\bibitem{papernot2016transferability}
N.~Papernot, P.~McDaniel, I.~Goodfellow \emph{et~al.}, ``Transferability in machine learning: from phenomena to black-box attacks using adversarial samples,'' \emph{arXiv preprint arXiv:1605.07277}, 2016.

\bibitem{chen2017zoo}
P.-Y. Chen, H.~Zhang, Y.~Sharma, J.~Yi, and C.-J. Hsieh, ``Zoo: Zeroth order optimization based black-box attacks to deep neural networks without training substitute models,'' in \emph{ACM workshop on artificial intelligence and security}, 2017.

\bibitem{yan2019subspace}
Z.~Yan, Y.~Guo, C.~Zhang \emph{et~al.}, ``Subspace attack: Exploiting promising subspaces for query-efficient black-box attacks,'' \emph{NeurIPS}, 2019.

\bibitem{chen2020hopskipjumpattack}
J.~Chen, M.~I. Jordan, M.~J. Wainwright \emph{et~al.}, ``Hopskipjumpattack: A query-efficient decision-based attack,'' in \emph{ieee symposium on security and privacy (sp)}, 2020.

\bibitem{dong2018boosting}
Y.~Dong, F.~Liao, T.~Pang, H.~Su, J.~Zhu, X.~Hu, and J.~Li, ``Boosting adversarial attacks with momentum,'' in \emph{CVPR}, 2018.

\bibitem{xie2019improving}
C.~Xie, Z.~Zhang, Y.~Zhou, S.~Bai, J.~Wang, Z.~Ren, and A.~L. Yuille, ``Improving transferability of adversarial examples with input diversity,'' in \emph{CVPR}, 2019.

\bibitem{cda}
M.~Naseer, S.~H. Khan, M.~H. Khan, F.~S. Khan, and F.~Porikli, ``Cross-domain transferability of adversarial perturbations,'' in \emph{NeurPIS}, 2019.

\bibitem{Zhang2022BIA}
Q.~Zhang, X.~Li, Y.~Chen, J.~Song, L.~Gao, Y.~He, and H.~Xue, ``Beyond imagenet attack: Towards crafting adversarial examples for black-box domains,'' in \emph{ICLR}, 2022.

\bibitem{ncf}
S.~Yuan, Q.~Zhang, L.~Gao, Y.~Cheng, and J.~Song, ``Natural color fool: Towards boosting black-box unrestricted attacks,'' in \emph{NeurIPS}, 2022.

\bibitem{li2020practical}
Q.~Li, Y.~Guo, H.~Chen \emph{et~al.}, ``Practical no-box adversarial attacks against dnns,'' in \emph{NeurIPS}, 2020.

\bibitem{ref_a24}
M.~D. Zeiler and R.~Fergus, ``Visualizing and understanding convolutional networks,'' in \emph{ECCV}, D.~J. Fleet, T.~Pajdla, B.~Schiele, and T.~Tuytelaars, Eds., 2014.

\bibitem{art}
A.~Oliva, ``The art of hybrid images: Two for the view of one,'' \emph{Art \& Perception}, vol.~1, no. 1-2, pp. 65--74, 2013.

\bibitem{sgm}
D.~Wu, Y.~Wang, S.~Xia, J.~Bailey, and X.~Ma, ``Skip connections matter: On the transferability of adversarial examples generated with resnets,'' in \emph{ICLR}, 2020.

\bibitem{ghost}
Y.~Li, S.~Bai, Y.~Zhou, C.~Xie, Z.~Zhang, and A.~L. Yuille, ``Learning transferable adversarial examples via ghost networks,'' in \emph{AAAI}, 2020.

\bibitem{sinifgsm}
J.~Lin, C.~Song, K.~He, L.~Wang, and J.~E. Hopcroft, ``Nesterov accelerated gradient and scale invariance for adversarial attacks,'' in \emph{ICLR}, 2020.

\bibitem{pifgsm}
L.~Gao, Q.~Zhang, J.~Song, X.~Liu, and H.~Shen, ``Patch-wise attack for fooling deep neural network,'' in \emph{ECCV}, 2020.

\bibitem{TMM1}
C.~Wan, F.~Huang, and X.~Zhao, ``Average gradient-based adversarial attack,'' \emph{IEEE Transactions on Multimedia}, vol.~25, pp. 9572--9585, 2023.

\bibitem{TMM2}
S.~Zhang, D.~Zuo, Y.~Yang, and X.~Zhang, ``A transferable adversarial belief attack with salient region perturbation restriction,'' \emph{IEEE Transactions on Multimedia}, vol.~25, pp. 4296--4306, 2023.

\bibitem{dong2019evading}
Y.~Dong, T.~Pang, H.~Su, and J.~Zhu, ``Evading defenses to transferable adversarial examples by translation-invariant attacks,'' in \emph{CVPR}, 2019.

\bibitem{long2023frequency}
Y.~Long, Q.~Zhang, B.~Zeng, L.~Gao, X.~Liu, J.~Zhang, and J.~Song, ``Frequency domain model augmentation for adversarial attack,'' in \emph{ECCV}, 2022.

\bibitem{dr}
Y.~Lu, Y.~Jia, J.~Wang, B.~Li, W.~Chai, L.~Carin, and S.~Velipasalar, ``Enhancing cross-task black-box transferability of adversarial examples with dispersion reduction,'' in \emph{CVPR}, 2020.

\bibitem{ssp}
M.~Naseer, S.~H. Khan, M.~Hayat, F.~S. Khan, and F.~Porikli, ``A self-supervised approach for adversarial robustness,'' in \emph{CVPR}, 2020.

\bibitem{jo2017measuring}
J.~Jo and Y.~Bengio, ``Measuring the tendency of cnns to learn surface statistical regularities,'' \emph{arXiv preprint arXiv:1711.11561}, 2017.

\bibitem{HF}
H.~Wang, X.~Wu, Z.~Huang, and E.~P. Xing, ``High-frequency component helps explain the generalization of convolutional neural networks,'' in \emph{CVPR}, 2020.

\bibitem{texture-bais}
R.~Geirhos, P.~Rubisch, C.~Michaelis, M.~Bethge, F.~A. Wichmann, and W.~Brendel, ``Imagenet-trained cnns are biased towards texture; increasing shape bias improves accuracy and robustness,'' in \emph{ICLR}, 2019.

\bibitem{aydemir2018effects}
A.~E. Aydemir, A.~Temizel, T.~T. Temizel \emph{et~al.}, ``The effects of jpeg and jpeg2000 compression on attacks using adversarial examples,'' \emph{arXiv preprint arXiv:1803.10418}, 2018.

\bibitem{das2018shield}
N.~Das, M.~Shanbhogue, S.-T. Chen, F.~Hohman, S.~Li, L.~Chen, M.~E. Kounavis, and D.~H. Chau, ``Shield: Fast, practical defense and vaccination for deep learning using jpeg compression,'' in \emph{KDD}, 2018.

\bibitem{liu2019feature}
C.~Liu and J.~JaJa, ``Feature prioritization and regularization improve standard accuracy and adversarial robustness,'' in \emph{IJCAI}, 2019.

\bibitem{featuredenoising}
C.~Xie, Y.~Wu, L.~van~der Maaten, A.~L. Yuille, and K.~He, ``Feature denoising for improving adversarial robustness,'' in \emph{CVPR}, 2019.

\bibitem{liu2019universal}
H.~Liu, R.~Ji, J.~Li, B.~Zhang, Y.~Gao, Y.~Wu, and F.~Huang, ``Universal adversarial perturbation via prior driven uncertainty approximation,'' in \emph{ICCV}, 2019.

\bibitem{wang2023towards}
Z.~Wang, H.~Yang, Y.~Feng, P.~Sun, H.~Guo, Z.~Zhang, and K.~Ren, ``Towards transferable targeted adversarial examples,'' in \emph{CVPR}, 2023.

\bibitem{zhou2020dast}
M.~Zhou, J.~Wu, Y.~Liu, S.~Liu, and C.~Zhu, ``Dast: Data-free substitute training for adversarial attacks,'' in \emph{CVPR}, 2020.

\bibitem{zhou2018transferable}
W.~Zhou, X.~Hou, Y.~Chen, M.~Tang, X.~Huang, X.~Gan, and Y.~Yang, ``Transferable adversarial perturbations,'' in \emph{ECCV}, 2018.

\bibitem{ganeshan2019fda}
A.~Ganeshan and R.~V. Babu, ``Fda: Feature disruptive attack,'' in \emph{ICCV}, 2019.

\bibitem{inkawhich2019feature}
N.~Inkawhich, W.~Wen, H.~H. Li, and Y.~Chen, ``Feature space perturbations yield more transferable adversarial examples,'' in \emph{CVPR}, 2019.

\bibitem{zhang2020understanding}
C.~Zhang, P.~Benz, T.~Imtiaz, and I.-S. Kweon, ``Understanding adversarial examples from the mutual influence of images and perturbations,'' in \emph{CVPR}, 2020.

\bibitem{li2020regional}
Y.~Li, S.~Bai, C.~Xie, Z.~Liao, X.~Shen, and A.~L. Yuille, ``Regional homogeneity: Towards learning transferable universal adversarial perturbations against defenses,'' in \emph{ECCV}, 2020.

\bibitem{ref_a33}
A.~M. Nguyen, J.~Yosinski, J.~Clune \emph{et~al.}, ``Deep neural networks are easily fooled: High confidence predictions for unrecognizable images,'' in \emph{CVPR}, 2015.

\bibitem{advesarialpatch}
T.~B. Brown, D.~Man{\'{e}}, A.~Roy, M.~Abadi, and J.~Gilmer, ``Adversarial patch,'' \emph{CoRR}, vol. abs/1712.09665, 2017.

\bibitem{eccvbias}
A.~Liu, J.~Wang, X.~Liu, B.~Cao, C.~Zhang, and H.~Yu, ``Bias-based universal adversarial patch attack for automatic check-out,'' in \emph{ECCV}, 2020.

\bibitem{zagoruyko2016wide}
S.~Zagoruyko and N.~Komodakis, ``Wide residual networks,'' in \emph{BMVC}, 2016.

\bibitem{hu2018squeeze}
J.~Hu, L.~Shen, G.~Sun \emph{et~al.}, ``Squeeze-and-excitation networks,'' in \emph{CVPR}, 2018.

\bibitem{pna}
C.~Liu, B.~Zoph, J.~Shlens, W.~Hua, L.~Li, L.~Fei{-}Fei, A.~L. Yuille, J.~Huang, and K.~Murphy, ``Progressive neural architecture search,'' in \emph{ECCV}, 2018.

\bibitem{shuff}
N.~Ma, X.~Zhang, H.~Zheng, and J.~Sun, ``Shufflenet {V2:} practical guidelines for efficient {CNN} architecture design,'' in \emph{ECCV}, V.~Ferrari, M.~Hebert, C.~Sminchisescu, and Y.~Weiss, Eds., 2018.

\bibitem{squee}
F.~N. Iandola, M.~W. Moskewicz, K.~Ashraf, S.~Han, W.~J. Dally, and K.~Keutzer, ``Squeezenet: Alexnet-level accuracy with 50x fewer parameters and {\textless}1mb model size,'' in \emph{ICLR}, 2017.

\bibitem{mob}
M.~Sandler, A.~G. Howard, M.~Zhu, A.~Zhmoginov, and L.~Chen, ``Mobilenetv2: Inverted residuals and linear bottlenecks,'' in \emph{CVPR}, 2018.

\bibitem{dcl}
Y.~Chen, Y.~Bai, W.~Zhang, and T.~Mei, ``Destruction and construction learning for fine-grained image recognition,'' in \emph{CVPR}, 2019.

\bibitem{imagenet}
O.~Russakovsky, J.~Deng, H.~Su, J.~Krause, S.~Satheesh, S.~Ma, Z.~Huang, A.~Karpathy, A.~Khosla, M.~S. Bernstein, A.~C. Berg, and F.~Li, ``Imagenet large scale visual recognition challenge,'' \emph{International Journal of Computer Vision}, 2015.

\bibitem{cub}
C.~Wah, S.~Branson, P.~Welinder, P.~Perona, and S.~Belongie, ``{The Caltech-UCSD Birds-200-2011 Dataset},'' California Institute of Technology, Tech. Rep., 2011.

\bibitem{car}
J.~Krause, J.~Deng, M.~Stark, and L.~Fei-Fei, ``Collecting a large-scale dataset of fine-grained cars,'' 2013.

\bibitem{air}
S.~Maji, E.~Rahtu, J.~Kannala, M.~B. Blaschko, and A.~Vedaldi, ``Fine-grained visual classification of aircraft,'' vol. abs/1306.5151, 2013.

\bibitem{kingma2014adam}
D.~P. Kingma and J.~Ba, ``Adam: A method for stochastic optimization,'' in \emph{ICLR}, 2015.

\bibitem{pifgsm++}
L.~Gao, Q.~Zhang, J.~Song, and H.~T. Shen, ``Patch-wise++ perturbation for adversarial targeted attacks,'' \emph{CoRR}, vol. abs/2012.15503, 2020.

\bibitem{huang2019enhancing}
Q.~Huang, I.~Katsman, H.~He, Z.~Gu, S.~Belongie, and S.-N. Lim, ``Enhancing adversarial example transferability with an intermediate level attack,'' in \emph{ICCV}, 2019.

\bibitem{brendel2017decision}
W.~Brendel, J.~Rauber, M.~Bethge \emph{et~al.}, ``Decision-based adversarial attacks: Reliable attacks against black-box machine learning models,'' in \emph{ICLR}, 2018.

\bibitem{maskrcnn}
K.~He, G.~Gkioxari, P.~Doll{\'{a}}r, and R.~B. Girshick, ``Mask {R-CNN},'' in \emph{ICCV}, 2017.

\bibitem{fasterrcnn}
S.~Ren, K.~He, R.~B. Girshick, and J.~Sun, ``Faster {R-CNN:} towards real-time object detection with region proposal networks,'' \emph{{IEEE} Trans. Pattern Anal. Mach. Intell.}, 2017.

\bibitem{retina}
T.~Lin, P.~Goyal, R.~B. Girshick, K.~He, and P.~Doll{\'{a}}r, ``Focal loss for dense object detection,'' in \emph{ICCV}, 2017.

\bibitem{fcn}
J.~Long, E.~Shelhamer, and T.~Darrell, ``Fully convolutional networks for semantic segmentation,'' in \emph{CVPR}, 2015.

\bibitem{deeplabv3}
L.~Chen, G.~Papandreou, F.~Schroff, and H.~Adam, ``Rethinking atrous convolution for semantic image segmentation,'' \emph{CoRR}, vol. abs/1706.05587, 2017.

\bibitem{ccnet}
Z.~Huang, X.~Wang, Y.~Wei, L.~Huang, H.~Shi, W.~Liu, and T.~S. Huang, ``Ccnet: Criss-cross attention for semantic segmentation,'' \emph{{IEEE} Trans. Pattern Anal. Mach. Intell.}, 2023.

\bibitem{eat}
F.~Tram{\`{e}}r, A.~Kurakin, N.~Papernot, I.~J. Goodfellow, D.~Boneh, and P.~D. McDaniel, ``Ensemble adversarial training: attacks and defenses,'' in \emph{ICLR}, 2018.

\bibitem{guo2018low}
C.~Guo, J.~S. Frank, and K.~Q. Weinberger, ``Low frequency adversarial perturbation,'' in \emph{UAI}, A.~Globerson and R.~Silva, Eds., 2019.

\bibitem{sharma2019on}
Y.~Sharma, G.~W. Ding, and M.~A. Brubaker, ``On the effectiveness of low frequency perturbations,'' in \emph{IJCAI}, S.~Kraus, Ed., 2019.

\bibitem{dziugaite2016study}
G.~K. Dziugaite, Z.~Ghahramani, and D.~M. Roy, ``A study of the effect of jpg compression on adversarial images,'' \emph{arXiv preprint arXiv:1608.00853}, 2016.

\end{thebibliography}

\clearpage

\setcounter{section}{0}

\newpage

\begin{figure*}[t]
  \centering
  \vspace*{2\baselineskip}
  \Large\bfseries Appendix for “Practical No-box Adversarial Attacks with Training-free Hybrid Image Transformation”
  \vspace*{2\baselineskip}
\end{figure*}
% \resumetocwriting
\renewcommand\thesection{\Alph{section}}

\section{Quantitative analysis about HFC and LFC}
\label{S.B}
\begin{figure}[h]
\centering
\includegraphics[height=4cm]{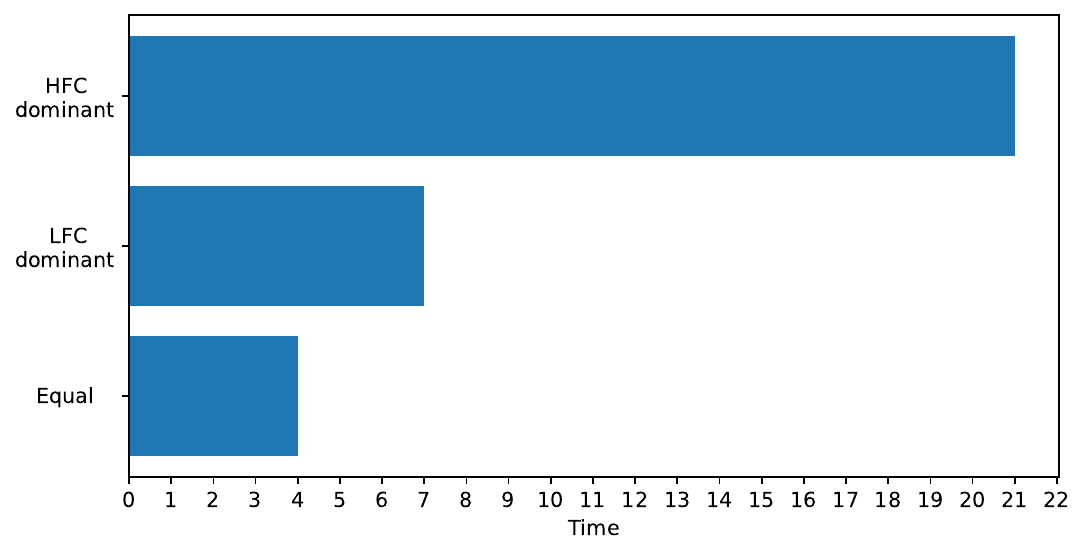}
\caption{We compare the average responses of HFC with LFC for each feature map. ``HFC dominant” means average responds of HFC is higher than LFC, and ``LFC dominant” is vice versa.}
\label{HFCvsLFC}
\vspace{-0.7cm}
\end{figure}
\begin{figure}[h]
\centering
\includegraphics[height=4cm]{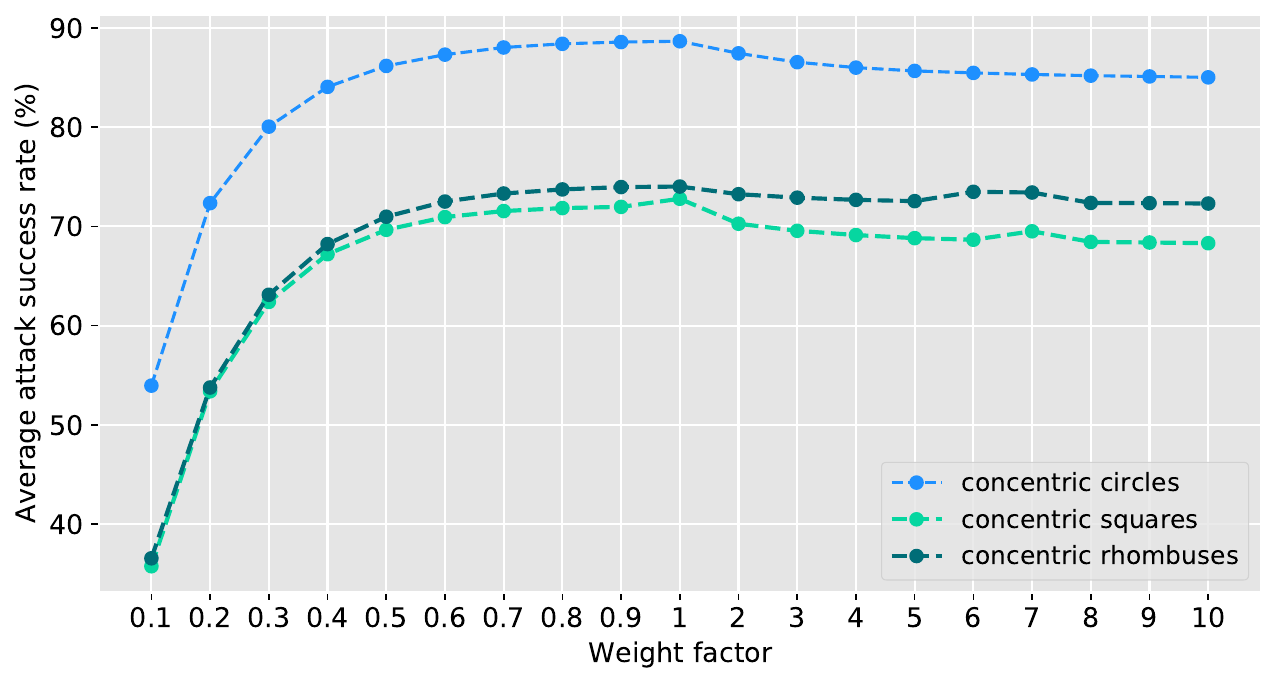}
\caption{The average attack success rates (\%) of different adversarial patches against normally trained models (for ImageNet) w.r.t. weight factors. The maximum perturbation $\varepsilon=16$.}
\label{weight}
\end{figure}
\begin{figure}[h]
\centering
\includegraphics[height=5cm]{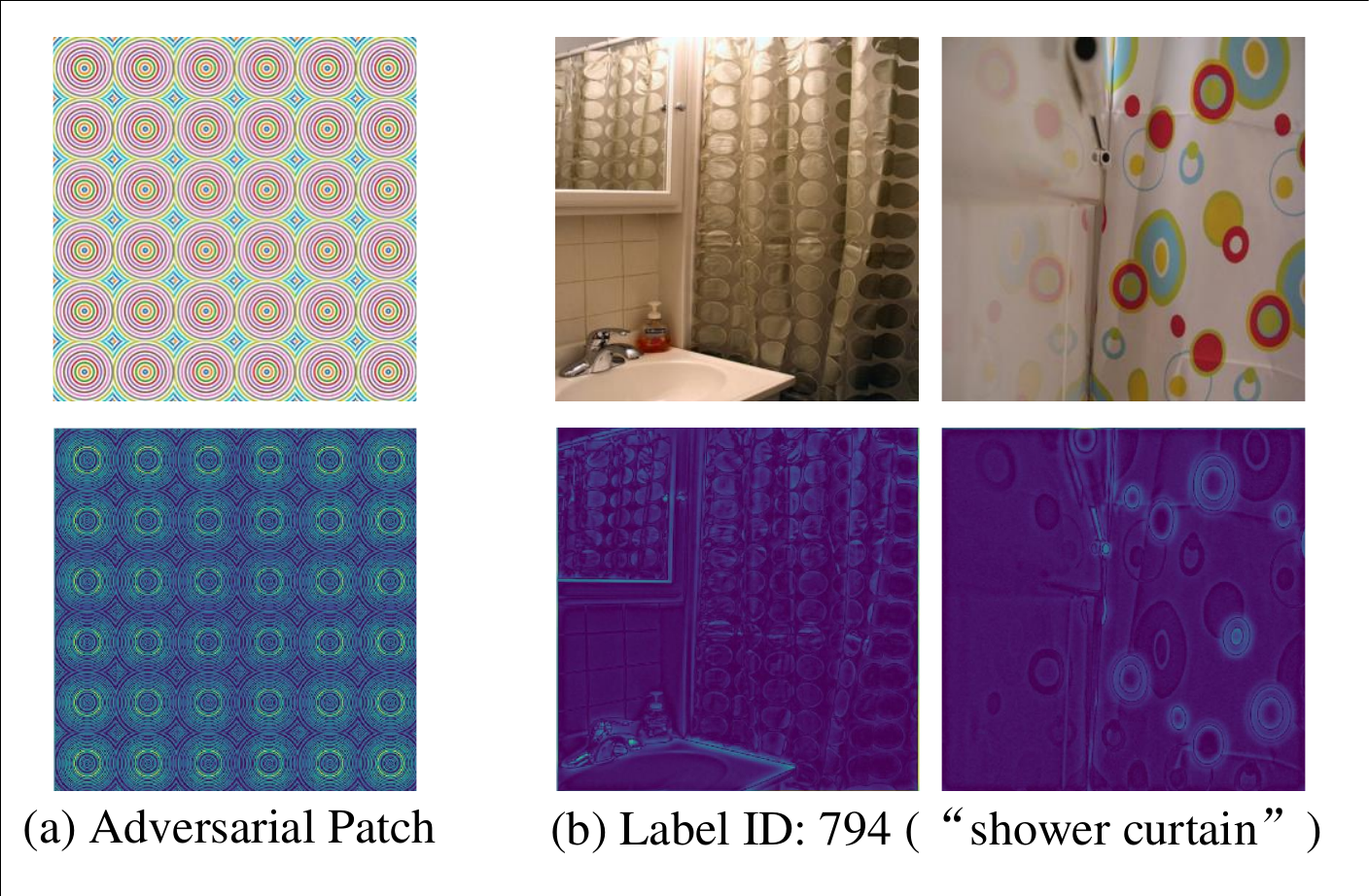}
\caption{We show (a) our adversarial patch and (b) some images which labeled as ``shower curtain" from ImageNet dataset. The bottom row is corresponding HFC extracted by Eq.~1.}
\label{794}
\vspace{-0.7cm}
\end{figure}
To quantitatively analyze whether HFC or LFC is dominant in the feature map of shallow layer, we conducted this experiment. Considering that the size of each shallow-layer feature map in Fig.~2(b) (in the main paper) is $147\time147$, we first resize Fig.~2(a) ($299\times299$) to $147\times147$, and denote resultant image by $\bm{x_r}$.
Then we get $\bm{x^H_r}$ (HFC of $\bm{x_r}$) by:
% \begin{figure*}[h]
% \centering
% \includegraphics[height=12cm]{nips_img/adv2022.pdf}
% \caption{Qualitative comparison for adversarial examples ($\varepsilon=16$) crafted by different methods.}
% \label{adv_cmp}
% \end{figure*}
% \begin{equation}
% \bm{x^H_r}=\bm{x_r}-\bm{x_r}*\bm{G}.
% \end{equation}
To quantitatively compare the response of HFC and LFC, we calculate the average response of each feature map $\phi(\bm{x})$ in low-frequency regions versus that in the other HFC regions. To that end, we generate two masks to distinguish these two regions. More specifically, the mask of high-frequency regions $\bm{M^H}$ can be written as:
\begin{equation}
\bm{M^H_{i,j}}=\begin{cases}
	1 ,&  |\bm{x^H_{r(i,j)}}| > \tau \\
	0 ,& else
\end{cases},
\end{equation}
where $\tau=20$ is pre-set threshold which applied to filter out low response. After getting $\bm{M^H}$, the mask of LFC $\bm{M^L}$ can be easy derivated:
\begin{equation}
\bm{M^L} = 1 - \bm{M^H}.
\end{equation}	
Therefore, the average response of HFC $a^H$ and the average response of LFC $a^L$ can be expressed as:
\begin{align}
\bm{a^H} = \frac{ \sum_{i,j}\bm{M^H}\odot\phi(\bm{x})}{ \sum_{i,j}\bm{M^H}},\\
\bm{a^L} = \frac{ \sum_{i,j}\bm{M^L}\odot\phi(\bm{x})}{ \sum_{i,j}\bm{M^L}}.
\end{align}
In this paper, if a feature map meets $\bm{a^H} > \bm{a^L}$, we call it ``HFC dominant”, otherwise we call it ``LFC dominant”.  
As demonstrate in Fig.~\ref{HFCvsLFC}, most feature maps are focus on HFC, and the ``HFC dominant” to ``LFC dominant” ratio is \textbf{3:1}.

\begin{figure}[h]
\centering
\includegraphics[height=3cm]{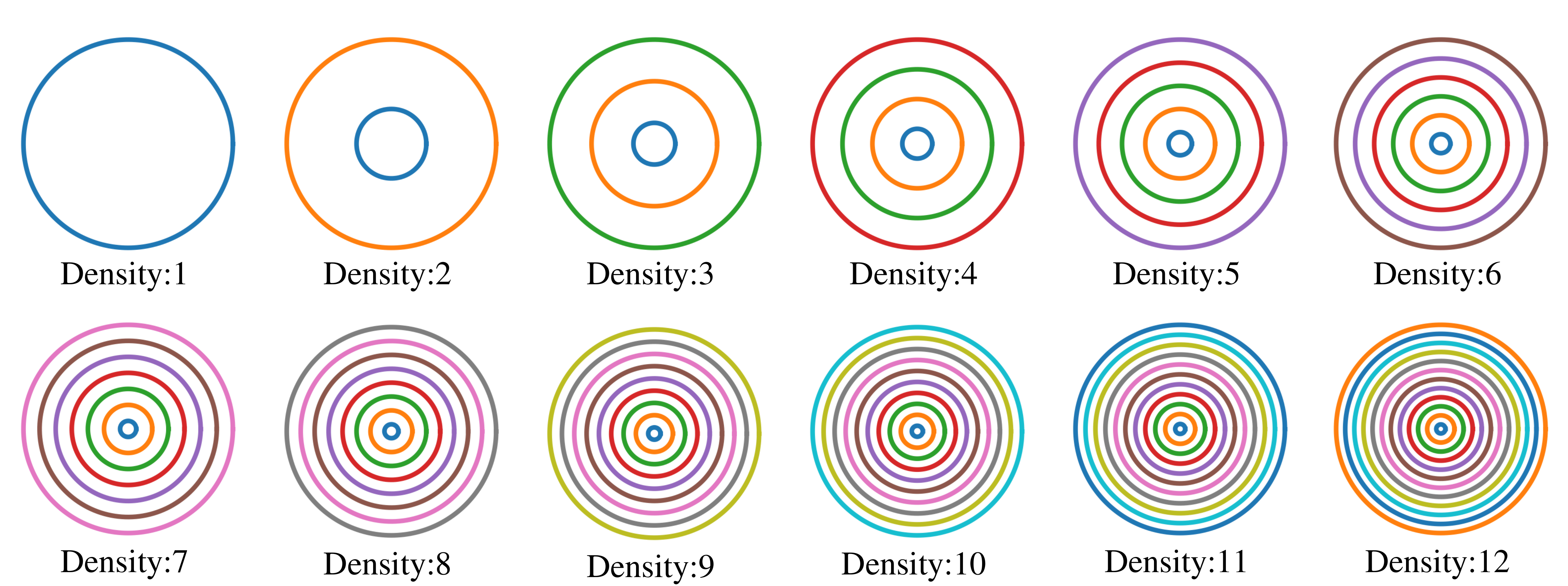}
\caption{We visualize our proto-patterns w.r.t. densities. Here we take concentric circles as an example. }
\label{d}
\vspace{-0.2cm}
\end{figure}

\begin{figure}[h]
\centering
\includegraphics[height=4cm]{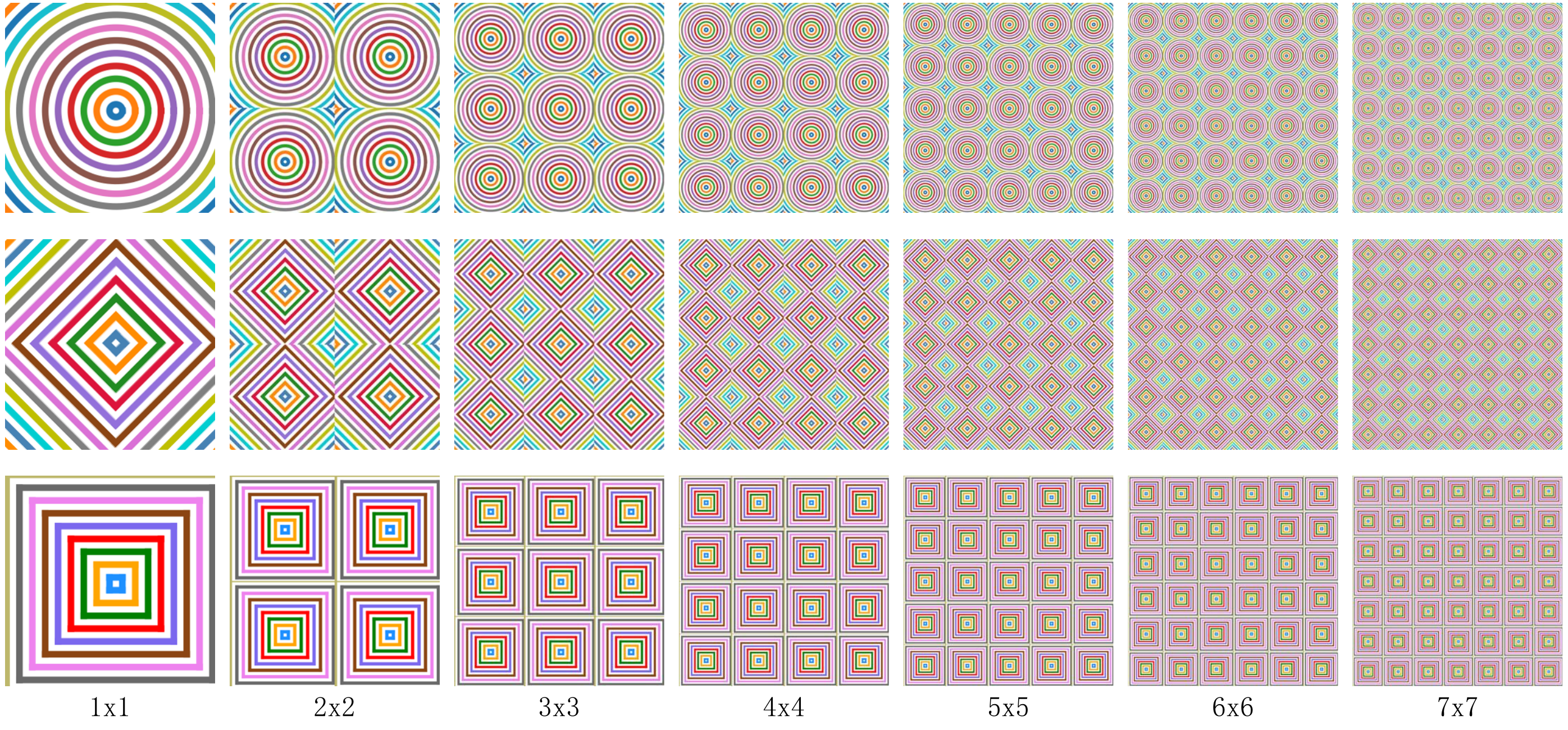}
\caption{We visualize our adversarial patches w.r.t. tile-schemes.}
\label{pattern}
\vspace{-0.5cm}
\end{figure}
\section{The Effect of Weight Factor $\lambda$}
\label{lambda_discuss}
In this section, we discuss the effect of different weight factors $\lambda$ on the experimental results. We tune $\lambda$ from 0.1 to 10, and the results are shown in Fig.~\ref{weight}. When $\lambda \leq 1$, the attack success rate increases rapidly at the beginning and then remains stable. However, further increasing $\lambda$ from $1$ to $10$ does not improve the performance. Actually, the success rates keep stable with a slight drop.

Apparently, a larger $\lambda$ leads to more perturbations (\textit{i.e.}, increase average perturbation of all pixels), and our reported results are a little inconsistent with linear assumption~\cite{goodfellow2014explaining}. It probably because our HIT is completely independent of any prior information (\textit{e.g.} the gradient of any model or data distribution). So it is not the larger the noise is, the farther the deviation from the true label will be. Besides, we notice that activation functions of these victim's models are all \textit{Relu}, which may be another reason for this phenomenon. More specifically, \textit{Relu} is defined as
\begin{equation}
	\mathit{Relu}(\bm{z}) = \begin{cases}
		0 ,&  \bm{z} < 0. \\
		\bm{z} ,& else.
	\end{cases}
\end{equation}
where $\bm{z}$ is the intermediate output before activation layer. If the intermediate adversarial perturbation is large enough, \textit{i.e.}, $\bm{\delta'} \leq -\bm{z}$, then $\mathit{Relu}(\bm{z}+\bm{\delta'})$ will return 0. But for a misclassification label $y^{adv} \neq y$, positive activation which is different from the original $\bm{z}$ may be more helpful than 0.

\section{Evaluation on Detection and Segmentation}

To make our experiment more solid, here we conduct experiments on other visual tasks, \textit{i.e.}, detection and segmentation. Specifically, we use MMDetection\footnote{\url{https://github.com/open-mmlab/mmdetection}} toolbox and the full COCO validation set\footnote{\url{https://cocodataset.org/\#home}} 
for the detection task and MMSegmentation\footnote{\url{https://github.com/open-mmlab/mmsegmentation}} toolbox and the full VOC2012 validation set\footnote{\url{http://host.robots.ox.ac.uk/pascal/VOC/voc2012/}} for the segmentation task. To avoid cherry-picking, we adopt the default setting of our HIT, \textit{i.e.}, $6\times 6$ Circle. 
The following Tab.~\ref{detection} shows the performance of our method. Notably, in the absence of any prior knowledge, our approach still significantly degrades the performance of existing models, which demonstrates the effectiveness of our HIT.
\begin{table}[h]

\centering
\resizebox{1\linewidth}{!}{
\begin{tabular}{ccccccccc|ccccc}
\toprule
\multicolumn{1}{c}{\multirow{2}{*}{Method}} & \multicolumn{2}{c}{MaskRCNN~\cite{maskrcnn}} & \multicolumn{2}{c}{FasterRCNN~\cite{fasterrcnn}} & \multicolumn{2}{c}{RetinaNet~\cite{retina}} & \multicolumn{2}{c|}{AVG.}  & \multicolumn{1}{c}{\multirow{2}{*}{FCN~\cite{fcn}}}   & \multicolumn{1}{c}{\multirow{2}{*}{DeepLabV3~\cite{deeplabv3}}} & \multicolumn{1}{c}{\multirow{2}{*}{CCNet~\cite{ccnet}}} & \multicolumn{1}{c}{\multirow{2}{*}{AVG.}}    \\
\multicolumn{1}{c}{}                        & AP            & AR           & AP             & AR            & AP            & AR            & AP           & AR           \\
\midrule
w/o attack                                  & 34.5          & 47.6         & 37.3           & 51.5          & 36.3          & 53.8          & 36.0  & 51.0  & 67.1 & 76.0      & 76.2 & 73.0 \\
\midrule
HIT(Circle w/ default tile-size)                                   & \textbf{8.5}           & \textbf{13.9}         & \textbf{9.4}            & \textbf{15.8}          & \textbf{8.8}           & \textbf{20.0}            & \textbf{8.9}          & \textbf{16.6} & \textbf{11.7} & \textbf{8.7}      & \textbf{11.9} & \textbf{10.7} \\
\midrule
\end{tabular}}
\caption{\textbf{Left}: The AP (average precision) and AR (average recall) of our HIT in object detection task on the full COCO valid set ($\varepsilon=16$). \textbf{Right}: The mloU of our HIT in segmentation task on the full VOC2012 valid set ($\varepsilon=16$).}
\label{detection}
\vspace{-0.5cm}
\end{table}

% \begin{table}[h]
% \caption{The mloU of our HIT in segmentation task on the full VOC2012 valid set ($\varepsilon=16$).}
% \centering
% \resizebox{0.5\linewidth}{!}{
% \begin{tabular}{ccccc}
% \toprule
% Method     & FCN   & DeepLabV3 & CCNet & AVG.    \\
% \midrule
% w/o attack & 67.1 & 76.0      & 76.2 & 73.0 \\
% \midrule
% HIT(Circle w/ default tile-size)  & \textbf{11.7} & \textbf{8.7}      & \textbf{11.9} & \textbf{10.7} \\
% \midrule
% \end{tabular}}
% \label{seg}
% \end{table}

\begin{table*}[h]

	\centering

	\resizebox{1.0\linewidth}{!}{
		\begin{tabular}{ccccccccccccccccccccc}
			\toprule
			\multirow{2}[4]{*}{} & \multicolumn{2}{c}{VGG19} & \multicolumn{2}{c}{Inc-v3} & \multicolumn{2}{c}{Res152} & \multicolumn{2}{c}{DenseNet} & \multicolumn{2}{c}{WRN} &\multicolumn{2}{c}{SENet} &\multicolumn{2}{c}{PNA} &\multicolumn{2}{c}{Shuffle} & \multicolumn{2}{c}{Squeeze} & \multicolumn{2}{c}{Mobile} \\
			\cmidrule{2-21}          & Label & \multicolumn{1}{l}{Ratio} & Label & \multicolumn{1}{l}{Ratio} & Label & \multicolumn{1}{l}{Ratio} & Label & \multicolumn{1}{l}{Ratio} & Label & \multicolumn{1}{l}{Ratio} & Label  & \multicolumn{1}{l}{Ratio} & Label & \multicolumn{1}{l}{Ratio} & Label & \multicolumn{1}{l}{Ratio} & Label& \multicolumn{1}{l}{Ratio} & Label & \multicolumn{1}{l}{Ratio}\\
			\midrule
Top-1 & 815 & 27.86 & \textbf{794} & \textbf{34.02} & \textbf{794} & \textbf{75.75} & 84 & 38.97 & 815 & 27.04 & \textbf{794} & \textbf{13.64} & \textbf{794} & \textbf{33.83} & 879 & 20.55 & 455 & 35.11 & \textbf{794} & \textbf{47.69} \\
Top-2 & 646 & 25.21 & 862 & 16.07 & 109 & 3.31 & \textbf{794} & \textbf{17.34} & 549 & 9.90 & 109 & 8.85 & 862 & 19.70 & 893 & 12.42 & \textbf{794} & \textbf{25.26} & 109 & 19.73 \\
Top-3 & 506 & 16.50 & 750 & 8.23 & 854 & 3.06 & 884 & 4.63 & 721 & 5.83 & 721 & 8.65 & 549 & 5.93 & 721 & 9.92 & 109 & 9.11 & 885 & 11.57 \\
Top-4 & \textbf{794} & \textbf{10.75} & 911 & 4.78 & 646 & 2.56 & 862 & 4.30 & 862 & 3.38 & 750 & 5.25 & 815 & 3.97 & \textbf{794} & \textbf{8.37} & 753 & 7.05 & 884 & 4.21 \\
Top-5 & 868 & 3.15 & 109 & 4.65 & 750 & 2.11 & 506 & 3.46 & 921 & 3.37 & 549 & 4.15 & 700 & 2.95 & 109 & 7.20 & 854 & 4.42 & 854 & 3.24\\
			\bottomrule
	\end{tabular}}%
	\caption{The top-5 label IDs that appear in classification results (range from 0 to 999) after HIT attack (tile-size is $50\times 50$, proto-pattern is concentric circle). The top row is victim's models, ratio (\%) represents the proportion of a specific prediction label to the total number of misclassified adversarial examples. The maximum perturbation $\varepsilon=16$.}
	\label{tab:target}%
\end{table*}%

\section{Discussion on Targeted Attack}
\label{app794}

\begin{figure*}[t]
\centering
\includegraphics[height=3.5cm]{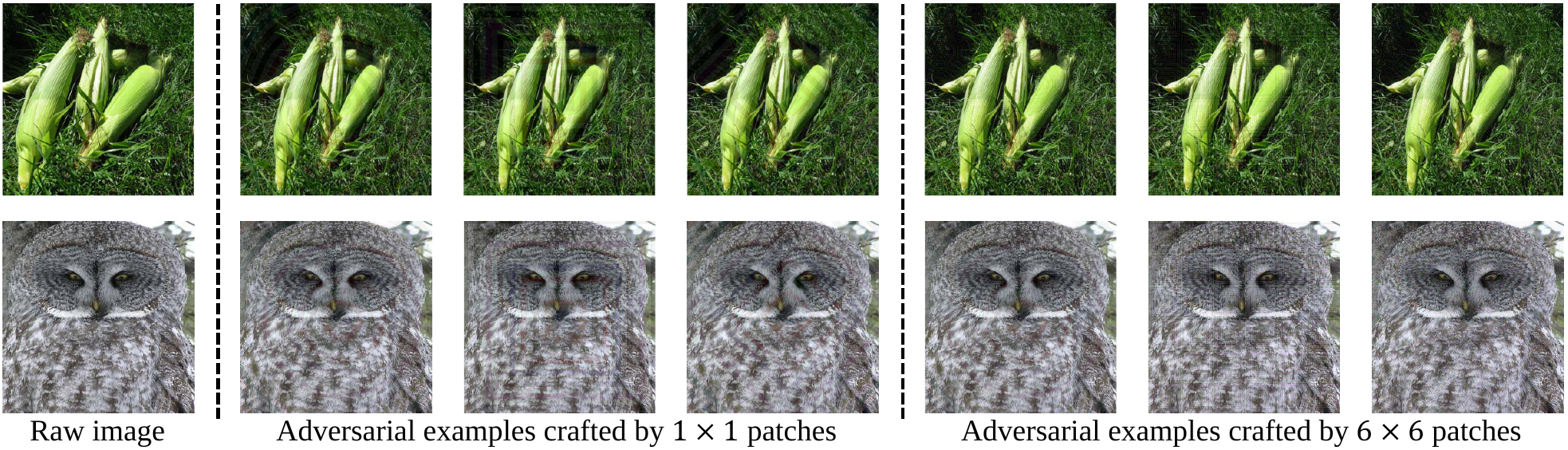}
\caption{The visualization for resultant adversarial examples ($\varepsilon=10$) w.r.t. tile-schemes.}
\label{fd_cmp}
\vspace{-0.5cm}
\end{figure*}
Although we do not explicitly force resultant adversarial examples to be misclassified as a specific targeted label, we observe that our HIT tends to implement a targeted attack due to the frequency domain operation and classification logic of DNNs. In Tab.~\ref{tab:target}, we report the top-5 prediction labels of our adversarial examples, which are crafted by $6\times 6$ concentric circle pattern. A first glance shows that almost all models tend to misclassify adversarial examples generated by our HIT as several specific labels, \textit{e.g.}, 794 (``shower curtain"). Furthermore, this phenomenon is more obvious for Mobile and Res152 whose ratio is up to \textbf{47.69\%} and \textbf{75.75\%} respectively.

To better understand this phenomenon, we show several clean images whose labels are ``shower curtain" and our adversarial patch in Fig.~\ref{794}. Interestingly, we observe that HFC of ``shower curtain" is somehow aligned with HFC of our adversarial patch, \textit{i.e.}, they all show similar certain repetitive circles. 
Therefore, our proposed perturbations might dominate overall features of the image and thus have the ability to achieve targeted attacks.
Since existing algorithms are not effective yet and simply replacing our adversarial patch with a clean targeted image cannot obtain an effective targeted attack, we will further study the selection and generation of adversarial patches, \textit{e.g.}, fusing shallow texture information of targeted distribution to guide resultant adversarial examples towards the targeted category.

\section{Visualization of Our Adversarial Patches}
\label{advesarialpatch}
In this section, we first visualize the concentric circle with respect to densities in Fig.~\ref{d}.
Here we control density from 1 to 12, \textit{e.g.}, ``2” denotes only two circles in the proto-pattern. With the increase of density, the distance between any two circles will also be reduced.
Then we list our adversarial patches with respect to tile-schemes in Fig.~\ref{pattern}. More specifically, we first crop $600\times 600 \times 3$ proto-patterns to $300\times 300 \times 3$ adversarial patches, then resize them into different tile-sizes (\textit{e.g.}, $150\times 150 \times 3$) and tile them back to $300\times 300 \times 3$, finally resize back to  $299\times 299\times 3$ to match the size of raw images. As we can see, if we decrease tile-size, distortion is inevitable.

\begin{figure}[h]
\centering
\includegraphics[height=3.2cm]{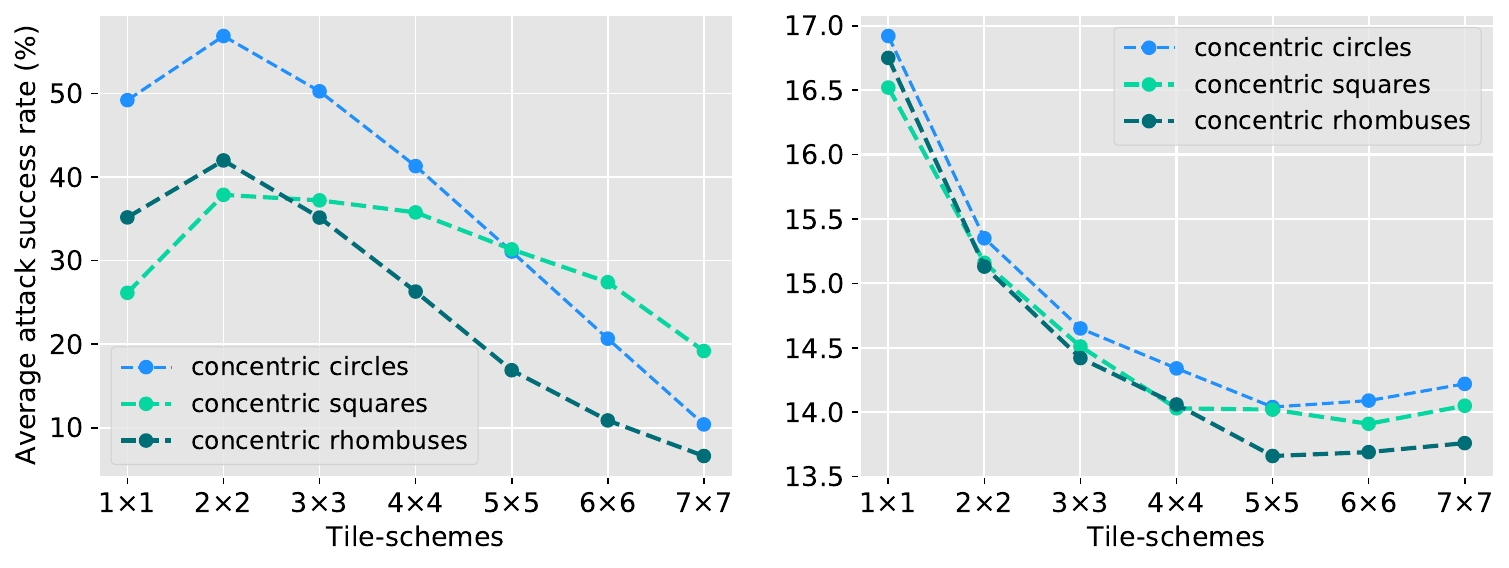}
\caption{The average attack success rates (\%) of different adversarial patches against EAT (left) and FD (right) w.r.t. tile-schemes. The maximum perturbation $\varepsilon=16$.}
\label{FD:FD}
\vspace{-0.5cm}
\end{figure}
\begin{figure}[t]
	\centering
% 	\vspace{-0.1cm}
	\includegraphics[width=0.8\linewidth]{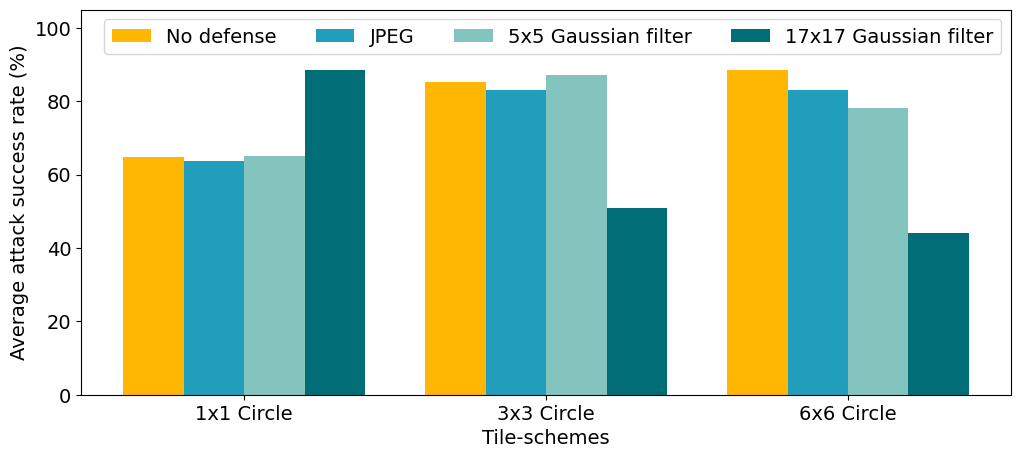}
% 	\vspace{-0.7cm}
	\caption{Average attack success rates (\%) against input transformation defenses on ten normally trained models introduced in our Sec. 4.}
	\label{fig1}
\end{figure}
\section{Attack against Adversarial Trained Defenses}
\label{fddiscussion}
In this section, we consider six additional well-known defense models, which including three ensemble adversarial training models (EAT)~\cite{eat}: Inc-v3$_{ens3}$, Inc-v3$_{ens4}$ and IncRes-v2$_{ens}$, and three feature denoising models (FD)~\cite{featuredenoising}: ResNet152 Baseline (Res152$_B$), ResNet152 Denoise (Res152$_D$), ResNeXt101 DenoiseAll (ResNeXt$_{DA}$), to demonstrate the effectiveness of our HIT.

As demonstrated in previous works\cite{guo2018low,sharma2019on}, low-frequency perturbations are more effective for attacking defense models. In our case, a smaller tile-scheme can generate a lower frequency perturbation. 
As shown in Fig.~\ref{fd_cmp}, the area of each regionally homogeneous (\textit{i.e.} continues) line in adversarial examples crafted by $1\times1$ patches is bigger than $6\times6$ ones. 
% Motivated by it, 
Therefore, we change tile-sizes to bigger ones (\textit{i.e.}, $150\times150$ for EAT and $300\times300$ for FD) and other parameters stay the same (see the ablation study in Fig.~\ref{FD:FD}). 
As observed in Tab.~\ref{eat}, our HIT is effective even for defense models. Notably, HIT can successfully attack Inc-v3$_{ens4}$ by \textbf{61.86\%}. Furthermore, for more robust FD, even crafting adversarial examples via an ensemble of VGG19, Inc-v3, Res152 and Dense, transfer-based S$^2$I-FGSM is still inferior to our HIT. This experimental result reveals that many defenses have not achieved real security, which are even vulnerable to training-free adversarial examples.

% \section{The Analysis of Tile-scheme for Defenses}
% \label{fddiscussion}
% In this section, we further consider six additional well-known defense models, which including three ensemble adversarial training models (EAT)~\cite{eat}: Inc-v3$_{ens3}$, Inc-v3$_{ens4}$ and IncRes-v2$_{ens}$, and three feature denoising models (FD)~\cite{featuredenoising}: ResNet152 Baseline (Res152$_B$), ResNet152 Denoise (Res152$_D$), ResNeXt101 DenoiseAll (ResNeXt$_{DA}$), to choose the best tile-scheme for fooling defense models.
% ###########################
\begin{table*}[t]

\centering
\resizebox{0.8\linewidth}{!}{
\begin{tabular}{ccp{2cm}<{\centering}p{2cm}<{\centering}p{1.5cm}<{\centering}p{1.5cm}<{\centering}p{1.5cm}<{\centering}p{1.5cm}<{\centering}p{1.5cm}<{\centering}}
	\toprule
	Model & Attack & Inc-v3$_{ens3}$ & Inc-v3$_{ens4}$ & IncRes$_{ens}$ & Res152$_B$   & Res152$_D$ & ResNeXt$_{DA}$   & AVG. \\
	\midrule
	-     & Raw images & 2.68  & 3.11  & 0.84  & 14.52  & 11.50  & 8.72  & 6.90  \\
	\midrule
	
    \multicolumn{1}{c}{\multirow{4}[0]{*}{\makecell[c]{VGG19, Inc-v3,\\Res152, Dense}}} 
	&TI-FGSM & 22.69  & 22.62  & 16.35  & 16.77  & 13.09  & 11.16  & 17.11  \\
	&DI$^2$-FGSM & 18.38  & 15.90  & 9.30  & 15.42  & 12.13  & 9.45  & 13.43  \\
	&PI-FGSM & 34.21  & 33.66  & 22.29  & 17.25  & 13.62  & 11.19  & 22.04  \\
% 	&PI-TI-DI$^2$-FGSM~\cite{pifgsm} & \textbf{70.04} & \textbf{69.43} & \textbf{56.37} & 18.47  & 14.58  & 12.33  & \textbf{40.20} \\
    &S$^2$I-FGSM & \textbf{71.44}&	\textbf{66.36}&	\textbf{50.98} 	&18.96 &15.58 		&12.37& 	\textbf{39.28} \\
	\midrule
	\multirow{1}[1]{*}{-} 
	&HIT (Ours) & 61.13  & 61.86  & 47.72  & \textbf{20.68} & \textbf{16.45} & \textbf{13.64} & 36.91  \\
	\bottomrule
\end{tabular}
}
\caption{The comparison of attack success rates (\%) on defense models between black-box attacks (adversarial examples are crafted via an ensemble of VGG19, Inc-v3, Res152 and Dense) and our no-box attacks with the maximum perturbation $\varepsilon=16$. The bold value denotes the best transferability.}
\label{eat}
\end{table*}

\begin{table*}[h]

\centering
\resizebox{0.8\linewidth}{!}{
\begin{tabular}{cccccccccccc}
\toprule
& VGG19 & Inc-v3 & Res152 & DenseNet & WRN & SENet & PNA & Shuffle & Squeeze & Mobile & AVG. \\
\midrule
1x1 w/o LF & 68.48 & 52.17 & 49.82 & 58.79 & 49.70 & 41.67 & 73.00 & 63.34 & 72.80 & 59.33 & 58.91 \\
1x1 w/ LF & \textbf{73.95} & \textbf{59.94} & \textbf{54.39} & \textbf{64.38} &\textbf{56.33} & \textbf{50.16} & \textbf{73.53} & \textbf{68.28} & \textbf{80.15} & \textbf{66.52} & \textbf{64.76} \\
\midrule
2x2 w/o LF & 91.32 & 72.74 & 67.21 & 76.79 & 68.48 & 54.61 & 83.54 & 78.27 & 88.46 & 88.31 & 76.97 \\
2x2 w/ LF & \textbf{92.89} & \textbf{77.33} & \textbf{71.29} & \textbf{80.42} & \textbf{72.80} & \textbf{60.17} & \textbf{85.21} & \textbf{82.45} & \textbf{91.54} & \textbf{90.32} & \textbf{80.44} \\
\midrule
3x3 w/o LF & 91.40 & 81.11 & 71.19 & 78.58 & 71.08 & 66.64 & 83.24 & 88.78 & 92.12 & 94.14 & 81.83 \\
3x3 w/ LF & \textbf{92.55} & \textbf{85.42} & \textbf{76.44} & \textbf{82.84} & \textbf{76.03} & \textbf{70.80} & \textbf{86.64} & \textbf{91.95} & \textbf{94.10} & \textbf{95.17} & \textbf{85.19} \\
\midrule
4x4 w/o LF & 92.91 & 83.89 & 76.02 & 83.78 & 69.55 & 63.68 & 82.18 & 91.62 & 94.82 & 95.89 & 83.43 \\
4x4 w/ LF & \textbf{93.97} & \textbf{88.28} & \textbf{82.69} & \textbf{88.34} & \textbf{76.85} & \textbf{69.27} & \textbf{86.18} & \textbf{94.93} & \textbf{96.65} & \textbf{96.81} & \textbf{87.40} \\
\midrule
5x5 w/o LF & 92.36 & 85.64 & 79.48 & 84.78 & 70.70 & 63.58 & 80.26 & 93.69 & 96.45 & 97.19 & 84.41 \\
5x5 w/ LF & \textbf{94.42} & \textbf{89.81} & \textbf{85.92} & \textbf{89.44} & \textbf{78.40} & \textbf{69.15} & \textbf{84.95} & \textbf{96.67} & \textbf{97.75} & \textbf{97.97} & \textbf{88.45} \\
\midrule
6x6 w/o LF & 92.99 & 85.86 & 81.49 & 83.41 & 71.82 & 65.92 & 76.43 & 95.00 & 97.36 & 96.96 & 84.72 \\
6x6 w/ LF & \textbf{94.75} & \textbf{90.37} & \textbf{87.62} & \textbf{88.80} & \textbf{79.26} & \textbf{70.31} & \textbf{82.12} & \textbf{97.34} & \textbf{98.31} & \textbf{97.81} & \textbf{88.67} \\
\bottomrule
\end{tabular}}
\caption{Average attack success rates of HIT (w/ Circle) w.r.t. tile-schemes. ``w/ LF" means adding our perturbations on LFC (i.e., reducing HFC beforehand) and ``w/o LF" means adding perturbations on benign samples. The maximum perturbation $\varepsilon=16$.}

\label{HITwoLF}
\end{table*}

\section{Attack against Input Pre-process Defenses}
In this section, we evaluate the performance of HIT (w/ Circle) on JPEG compression~\cite{dziugaite2016study} and Gaussian filter defenses, and Fig.~\ref{fig1} illustrates our results. We observe that neither JPEG compression (quality factor is 75) nor $5\times 5$ Gaussian filter can effectively defend against HIT-based adversarial examples. It is because our HFC is extracted by a $17\times 17$ Gaussian filter, whose cutoff frequency is relatively low. So most components in our HFC cannot be removed by above quality-preserving defenses.
To suppress our HFC, a larger Gaussian filter (e.g., $17\times 17$) is necessary. However, it only works for big tile-schemes, e.g., $6\times 6$. 
It is because fine (or thin) regionally homogeneous HFC is easier to filter, while coarse one is more difficult.
It can also help explain why $17\times 17$ Gaussian filter instead boosts HIT ($1\times 1$ tile-scheme): It cannot effectively filter out coarse regionally homogeneous perturbations but inevitably filter out many important details of an image (i.e., seriously degrade image quality), making it more difficult for DNNs to classify accurately. Thus, naive input transformation, such as JPEG compression or Gaussian filter, cannot defend against HIT.

\begin{figure*}[h]
\centering
\includegraphics[height=3cm]{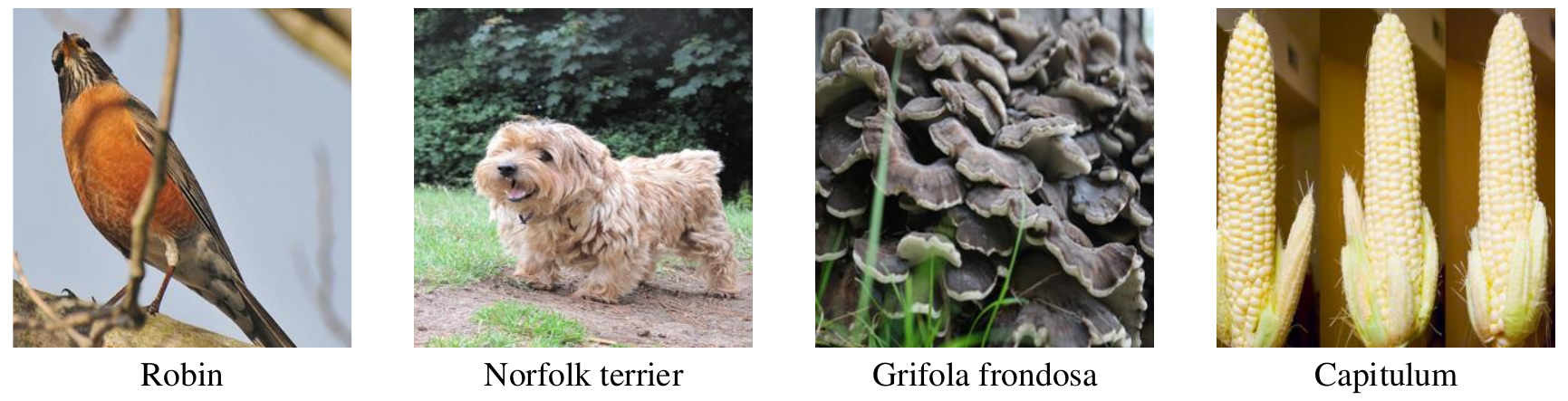}
\caption{We randomly select four raw images from ImageNet dataset~\cite{imagenet} to replace our adversarial patches, then test the attack performance by our HIT.}
\label{rawimg}
\end{figure*}

\begin{table*}[h]

\centering
\resizebox{1\linewidth}{!}{
\begin{tabular}{cccccccccccc}
\toprule
Attack & VGG19 & Inc-v3 & Res152 & Dense & WRN & SENet & PNA   & Shuffle & Squeeze & Mobile & AVG. \\
\midrule
HIT w/ Robin & 27.37  & 19.78  & 16.62  & 18.72  & 17.07  & 11.73  & 25.70  & 37.34  & 48.77  & 33.47  & 25.66  \\
HIT w/ Norfolk terrier & 33.01  & 28.06  & 21.87  & 27.94  & 24.88  & 19.39  & 34.57  & 42.70  & 59.34  & 39.45  & 33.12  \\
HIT w/ Grifola frondosa & 37.93  & 29.25  & 25.38  & 33.02  & 28.36  & 19.78  & 41.77  & 46.10  & 61.30  & 43.17  & 36.61  \\
HIT w/ Capitulum & 46.54  & 26.82  & 29.00  & 37.18  & 30.49  & 30.14  & 41.77  & 41.00  & 55.78  & 51.36  & 39.01  \\
\bottomrule
\end{tabular}}%
\caption{The comparison of attack success rates (\%) w.r.t. raw images. The maximum perturbation $\varepsilon=16$.}%
\label{rawtable}
\end{table*}%

\section{Raw Image for Attack}
\label{raw_img}
To highlight the effectiveness of our design for adversarial patches, here we conduct an experiment where raw images (shown in Fig.~\ref{rawimg}) serve as ``adversarial patch", \textit{i.e.}, utilizing HFC of these raw images to manipulate adversarial examples. 
However, even leveraging HFC of texture-rich raw images (\textit{e.g.}, ``Grifola frondosa" and ``Capitulum") cannot achieve a good result. As demonstrated in Tab.~\ref{rawtable}, the average attack success rates are all less than 40\%. By contrast, our well-designed adversarial patches can easily achieve a success rate of \textbf{88.67\%}(see Tab.~3 in main paper), which demonstrates the effectiveness of our design.

\section{The Effect of Removing HFC of Benign Samples}
\label{HIT_LF}
In this section, we show the effect of removing HFC of benign samples before adding noisy HFC. As demonstrated in Tab.~\ref{HITwoLF}, adding our perturbations on LFC (instead of benign samples) can make resultant adversarial examples fool target models more easily. 
% \clearpage

% \bibliography{main}

\vfill

\end{document}